\def\paperlanguage{} %% English

\pdfoutput=1 % for arxiv

\documentclass[letterpaper, 10 pt, conference]{ieeeconf}  % Comment this line out if you need a4paper

\usepackage{bm}
\usepackage{cite}
\usepackage{flushend}
\usepackage{url}
\usepackage{here}
\include{preamble}

\usepackage{booktabs}

\IEEEoverridecommandlockouts                              % This command is only needed if
\title{\LARGE \textbf
  {
    \switchlanguage%
    {%
    M3D-skin: Multi-material 3D-printed Tactile Sensor with Hierarchical Infill Structures for Pressure Sensing
    }%
    {%
      積層3Dプリントのインフィルパターンをセンサ原理に活用した、
      任意形状で作成可能な触覚センサ
    }%
  }
}

\author{Shunnosuke Yoshimura$^{1}$, Kento Kawaharazuka$^{1,2}$, and Kei Okada$^{1}$% <-this % stops a space
  \thanks{$^{1}$ The authors are with the Department of Mechano-Informatics, Graduate School of Information Science and Technology, The University of Tokyo, 7-3-1 Hongo, Bunkyo-ku, Tokyo, 113-8656, Japan.
    {\texttt\small [yoshimura, kawaharazuka, okada]@jsk.t.u-tokyo.ac.jp}
  }
  \thanks{$^{2}$ The author is with the AI Center, Graduate School of Information Science and Technology, The University of Tokyo, Japan.}
}
\begin{document}

\maketitle
\thispagestyle{empty}
\pagestyle{empty}

%%%%%%%%%%%%%%%%%%%%%%%%%%%%%%%%%%%%%%%%%%%%%%%%%%%%%%%%%%%%%%%%%%%%%%%%%%%%%%%%
\begin{abstract}
  \switchlanguage%
  {%
  Tactile sensors have a wide range of applications, from utilization in robotic grippers to human motion
  measurement. If tactile sensors could be fabricated and integrated more easily, their applicability would further
  expand.
  In this study, we propose a tactile sensor—M3D-skin—that can be easily fabricated with high versatility
  by leveraging the infill
  patterns of a multi-material 
  fused deposition modeling (FDM) 3D printer as the sensing principle. This method employs conductive and
  non-conductive flexible filaments to create a hierarchical structure with a specific infill pattern.
  The flexible hierarchical structure deforms under pressure, leading to a change in electrical resistance, enabling the
  acquisition of tactile information. 
  We measure the changes in characteristics of the proposed tactile sensor caused by modifications to the hierarchical structure.
  Additionally, we demonstrate the fabrication and use of a multi-tile
  sensor.
  Furthermore, as applications, we implement motion pattern measurement on the sole of a foot, integration with a robotic
  hand, and tactile-based robotic operations. Through these experiments, we validate the effectiveness of the proposed
  tactile sensor.
  }%
  {%
  ロボットの手先における触覚利用から、人体の運動計測まで、触覚センサは広い応用可能性を有する。
  触覚センサを、より様々な形状で、かつ簡単に作成することができれば、これらの応用性はより広がる。
  そこで、本研究では、積層型3Dプリンタのインフィルパターンをセンサ原理に活用した、
  任意形状で簡単に作成可能な触覚センサを提案する。
  この手法においては、導電性および非導電性の柔軟なフィラメントを用い、
  特定のインフィルパターンによる階層構造を作成する。
  この柔軟な改装構造は圧力を受けると変形して抵抗値が変化し、
  触覚情報を取得することが可能である。
  提案した触覚センサについて、
  階層構造の変更による特性変化を評価した後に、
  複数タイルのセンサの作成と利用が可能であることを示す。
  さらに、応用として、足裏における圧力パターン変化測定、
  ロボットハンドとの一体成型および触覚利用動作を実現する。
  これらの実験を通し、提案した触覚センサの有用性を示す。
  }%
\end{abstract}

\section{Introduction}\label{sec:introduction}

\begin{figure}[t]
  \centering
  \includegraphics[width=1.0\columnwidth]{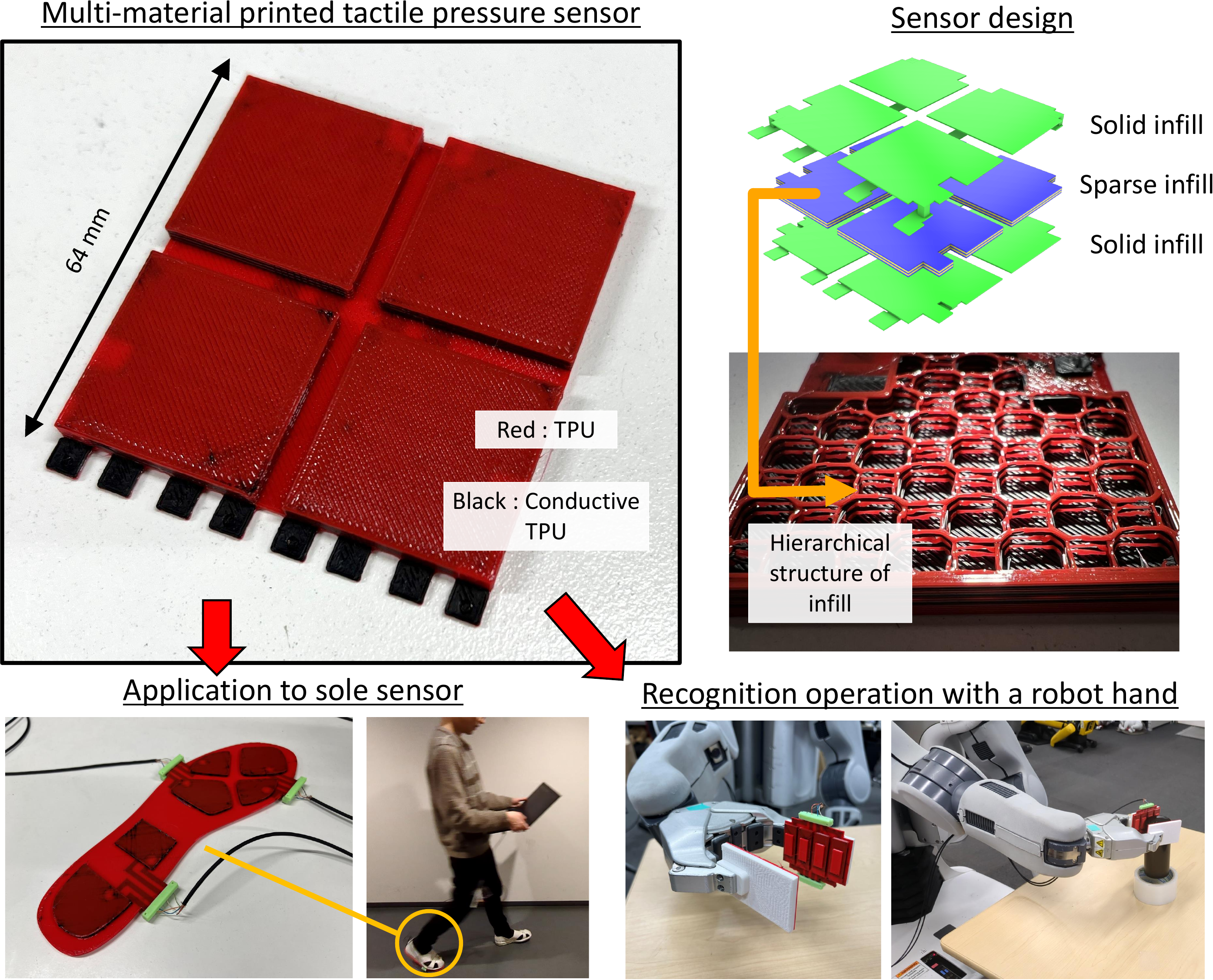}
  \caption{
    % 本研究の全体像を示す。
    % 上部には、3Dプリントされた触覚センサと、
    % その構造および内部のインフィルを示す。
    % 下部には、センサの応用例を示す。
    Overview of this study.
    The upper section illustrates the 3D-printed tactile sensor, M3D-skin, including its structure and internal infill pattern.
    The lower section presents application examples of the sensor.
  }
  \label{overview}
  \vspace{-3mm}
\end{figure}

\switchlanguage%
{%
Tactile sensors are widely utilized across various fields, ranging from control and recognition tasks in robotic object
grasping \cite{5771603, yuan2017gelsight} to human motion analysis in wearable devices \cite{yeo2016wearable,
titianova2004footprint}. In these applications, improving the ease of integration and enabling rapid and simple
development of tactile sensors would enhance their usability and expand their range of applications.

The primary sensing principles for tactile sensors include piezoresistive, capacitive, and piezoelectric methods
\cite{D2MH00892K, li2021synergy, 5969151, zhang2024kirigami}. However, these approaches often require specialized
fabrication techniques and may present challenges in terms of shape customization, scalability, and adaptability. To
address these issues, research has explored tactile sensors fabricated using rubber molding and 3D printing
\cite{james2021tactile, 10418152, guo20173d, tang20203d, kim2019highly, 8233949, massaroni2024fully}. In particular, if
a widely accessible sensor could be created using fused deposition modeling (FDM) 3D printing and general-purpose
materials, it would significantly contribute to the widespread adoption of tactile sensing technologies.

In this study, we propose a tactile sensor, M3D-skin, which utilizes the infill patterns of FDM 3D printers.
An overview of the concept is shown in \figref{overview}. This approach involves alternating layers of
conductive and non-conductive flexible filaments printed with a specific infill pattern, forming a hierarchical
structure. The deformation of this hierarchical structure under applied pressure induces a change in electrical
resistance, allowing for the acquisition of tactile information. Additionally, an embedded wiring layer enables the
creation and use of arbitrarily shaped, multi-tile sensors.

% In our experiments, we first evaluate how different hierarchical structures affect sensor characteristics, demonstrating
% their properties and tunability. Next, we fabricate and validate multi-tile sensor arrays. Finally, as practical
% applications, we implement motion pattern detection on the sole of a foot and integrate the sensor into a robotic hand
% for tactile-based operations. These experiments demonstrate that the proposed sensor can be easily and cost-effectively
% fabricated using FDM 3D printing while maintaining flexibility in shape, multi-tile configuration, and excellent
% integration and applicability.

In our experiments, we first measure the sensor's characteristics and evaluate how different hierarchical structures 
affect its performance, demonstrating their properties and adjustability. Next, we fabricate multi-tile sensors to verify 
their usability. Finally, as applications of this sensor, we implement motion pattern detection on the sole of a foot and 
integrate it into a robotic hand for tactile-based operations.
These experiments demonstrate that the proposed sensor can be easily and cost-effectively fabricated using an FDM 3D printer 
while enabling arbitrary shapes and multi-tile configurations, as well as excellent integration and applicability.
}%
{%
触覚センサは、ロボットの物体把持における制御や認識動作への利用\cite{5771603, yuan2017gelsight}から、
ウェアラブルデバイスにおける人体の運動解析\cite{yeo2016wearable, titianova2004footprint}まで、
広い分野で活用される。
これらの分野において、デバイスにより組み込みやすく、簡単かつ高速に触覚センサを開発できれば、
その応用範囲や利便性はより向上する。
触覚センサの原理においては、電気抵抗式、静電容量式、圧電式など
\cite{D2MH00892K, li2021synergy, 5969151, zhang2024kirigami}
が主流であるものの、
制作には専門的な技術が必要であり、
形状やカスタマイズ性、スケーラビリティにおいて問題が生じうる。
そこで、ゴム成形や3Dプリンタを用いた
触覚センサの研究
\cite{james2021tactile, 10418152, guo20173d, tang20203d, kim2019highly, 8233949, massaroni2024fully}
も行われている。
特に、近年急速に利便性が向上している積層型3Dプリンタと汎用素材を用いて
触覚センサを作成できれば、大衆的なセンサ技術としての普及が期待される。

そこで、本研究では、積層型3Dプリンタのインフィルパターンをセンサ原理に活用した、
触覚センサの設計を提案する。
この概要を\figref{overview}に示す。
この手法においては、特定のインフィルパターンで印刷された、
導電性および非導電性の柔軟なフィラメントの層を交互に積層して、
階層構造を作成する。
この階層構造は圧力を受けると変形して抵抗値が変化するために、
触覚情報を取得できる。
また、センサの内部に配線層を設け、
任意形状かつ複数タイルのセンサを作成し、利用することが可能である。
本研究の実験においては、まず、異なる階層構造がセンサ特性に与える影響を測定し、
その特性と調整可能性を示す。
次に、複数タイルのセンサを作成し、利用可能であることを示す。
最後に、このセンサの応用として、
足裏における運動パターン変化測定、
ロボットハンドとの一体成型および触覚利用動作を実現する。
これらの実験を通し、このセンサが、積層型3Dプリンタによって安価かつ簡単に制作可能で
ありながら、任意形状、複数タイルの触覚センサとして利用でき、
人体の動作測定やロボットハンドへの組み込みにおいて有用であることを示す。
}%

\subsection{Related Work}
\switchlanguage%
{%
The primary sensing principles for tactile sensors include piezoresistive, capacitive, and piezoelectric methods. Various
hybrid and novel sensor designs have been proposed, such as a capacitive–piezoresistive hybrid pressure sensor
\cite{D2MH00892K} and an electrically resistive pressure sensor utilizing a porous and microstructured piezoresistive
material \cite{li2021synergy}. Additionally, a high-sensitivity MEMS capacitive pressure sensor that separates pressure
sensing and capacitance measurement \cite{5969151} and a piezoelectric sensor leveraging a kirigami structure
\cite{zhang2024kirigami} have been developed.
Furthermore, new sensing principles using novel materials have been explored, 
such as ultrathin gold nanowires \cite{gong2014wearable} and optical fiber pressure sensors based on 
polarization-maintaining photonic crystal fiber \cite{fu2008pressure}.

To develop more easily integrable tactile sensors, research has explored fabrication using rubber molding and 3D
printing. A PolyJet-based 3D-printed integrated robotic hand and sensor system \cite{james2021tactile} enables pressure
distribution measurement through image recognition. Other approaches include pressure sensors combining Hall sensors
with printed structures \cite{10418152}, stretchable force sensors printed with silicon-based materials
\cite{guo20173d}, flexible pressure sensors based on piezoresistive composites fabricated with custom FDM printers
\cite{tang20203d}, and highly sensitive pressure sensors utilizing printed structures combined with elastomers
\cite{kim2019highly}.

% For simpler, lower-cost, and shape-customizable fabrication, an integrated sensor using commercial FDM 3D
% printers and readily available filaments is ideal. Previous research on FDM 3D-printed tactile sensors includes a
% method that creates capacitors through conductive filament layering, using capacitance changes for pressure sensing
% \cite{8233949}. However, this approach suffers from small capacitance changes in response to applied force and
% vulnerability to external disturbances. Another approach employs deformable protrusions on the sensor surface for
% pressure sensing \cite{massaroni2024fully}, but it requires large protrusions, limiting shape and size flexibility.

For simpler, lower-cost, and shape-customizable fabrication, an integrated sensor using commercial FDM 3D printers and
readily available filaments is ideal. Previous research includes capacitive sensors formed by layering conductive
filaments \cite{8233949}, but they exhibit small capacitance changes.
Another method uses deformable protrusions for pressure sensing \cite{massaroni2024fully}, but the required large protrusions
limit shape and size flexibility.

A key challenge in FDM 3D-printed sensor fabrication is that widely available, low-cost conductive and flexible
filaments are stiffer than rubber, leading to minimal deformation under pressure. 
% This limitation
% results in small capacitance changes in the capacitor-based approach \cite{8233949} and necessitates large
% pressure-sensitive protrusions in \cite{massaroni2024fully}. 
This results in small capacitance changes \cite{8233949} and the need for large protrusions \cite{massaroni2024fully}.
In contrast, this study proposes utilizing infill patterns,
a unique feature of FDM 3D printing, as the pressure-sensitive structure. This approach enables significant resistance
changes even with small deformations, overcoming these limitations.
}%
{%
触覚センサの原理は、電気抵抗式、静電容量式、圧電式などが主流である。
静電容量–ピエゾ抵抗
ハイブリッド型の圧力センサ\cite{D2MH00892K}
や、
A porous and microstructure piezoresistive materialを活用した、
電気抵抗式圧力センサ
\cite{li2021synergy}
が提案されている。
% 圧力感知と容量測定の分離による
高感度のMEMS静電容量式圧力センサ
\cite{5969151}
の制作や、
Kirigami構造を活用した圧電式センサ\cite{zhang2024kirigami}
といった特定構造の利用も行われている。
また、新たな素材によるセンサ原理開発として、
Ultrathin gold nanowiresの利用
\cite{gong2014wearable}
や、
polarization-maintaining photonic crystal fiberによる
光ファイバー圧力センサ
\cite{fu2008pressure}
% といった、既存原理と異なるセンサ
も提案されている。
これらのセンサは性能が優れるものの、制作には専門的な技術が必要であり、
ロボットや人体計測への組み込みにおいては、形状やカスタマイズ性
において問題が生じうる。

そこで、より組み込みやすい触覚センサについて、ゴム成形や
3Dプリント造形によるセンサの研究が行われている。
Polyjet方式\cite{polyjet}の3Dプリンタを利用した、一体成型ロボットハンドセンサ
\cite{james2021tactile}
では、
画像認識による圧力分布測定を実現している。
ホールセンサとプリント構造の組み合わせによる圧力センサ\cite{10418152}、
シリコン材料のプリントによるstretchableな力センサ\cite{guo20173d}、
自家製積層プリンタによる
Piezoresistive composite-based flexible pressure sensors\cite{tang20203d},
プリント構造とエラストマーの組み合わせによる高い感度の圧力センサ\cite{kim2019highly}
などが存在する。

これらと比べ、さらに簡単かつ安価に、任意形状で作成するためには、
市販の積層型3Dプリンタおよび入手性の高いフィラメントを利用した、
一体成型可能なセンサが望ましい。
積層型3Dプリンタを用いた触覚センサの研究として、
導電性フィラメントの積層によりキャパシタを作成し、
その容量変化を利用して圧力センサを作成する手法\cite{8233949}は、
与えた力に対する容量の変化が小さく、外乱に弱い。
変形可能な突起形状を活用した
センサ\cite{massaroni2024fully}
では、センサ表面に感圧のための大きな突起形状が必要であり、形状やサイズが限定される。

積層型3Dプリンタによるセンサ作成時に課題となるのが、
安価で入手しやすく、かつ印刷安定性の高い導電性フィラメントや柔軟フィラメントは、
ゴム素材と比べて固く、圧力を受けた際に小さな変形しか生じないことである。
ゆえに、\cite{8233949}のキャパシタ方式では容量変化が小さく、
\cite{massaroni2024fully}では大きな突起が感圧部に必要になっていた。
一方、本研究では、感圧のための構造に、
積層型3Dプリンタに特有のインフィルパターンを利用することで、
小さな変形でも十分な抵抗値変化を得ることができる。
}%

\section{Method}\label{sec:method}
\subsection{Sensor Structure}
\switchlanguage%
{%
The overall structure of M3D-skin is shown in \figref{sensor_structure}.
It is fabricated using a FDM 3D printer with conductive and non-conductive TPU (Thermoplastic Polyurethane)
filaments.
This sensor consists of three structural components: a sensor layer for pressure sensing, wiring layers for
consolidating the wiring, and cover layers for surface protection.

\begin{figure}[t] 
  \centering 
  \includegraphics[width=1.0\columnwidth]{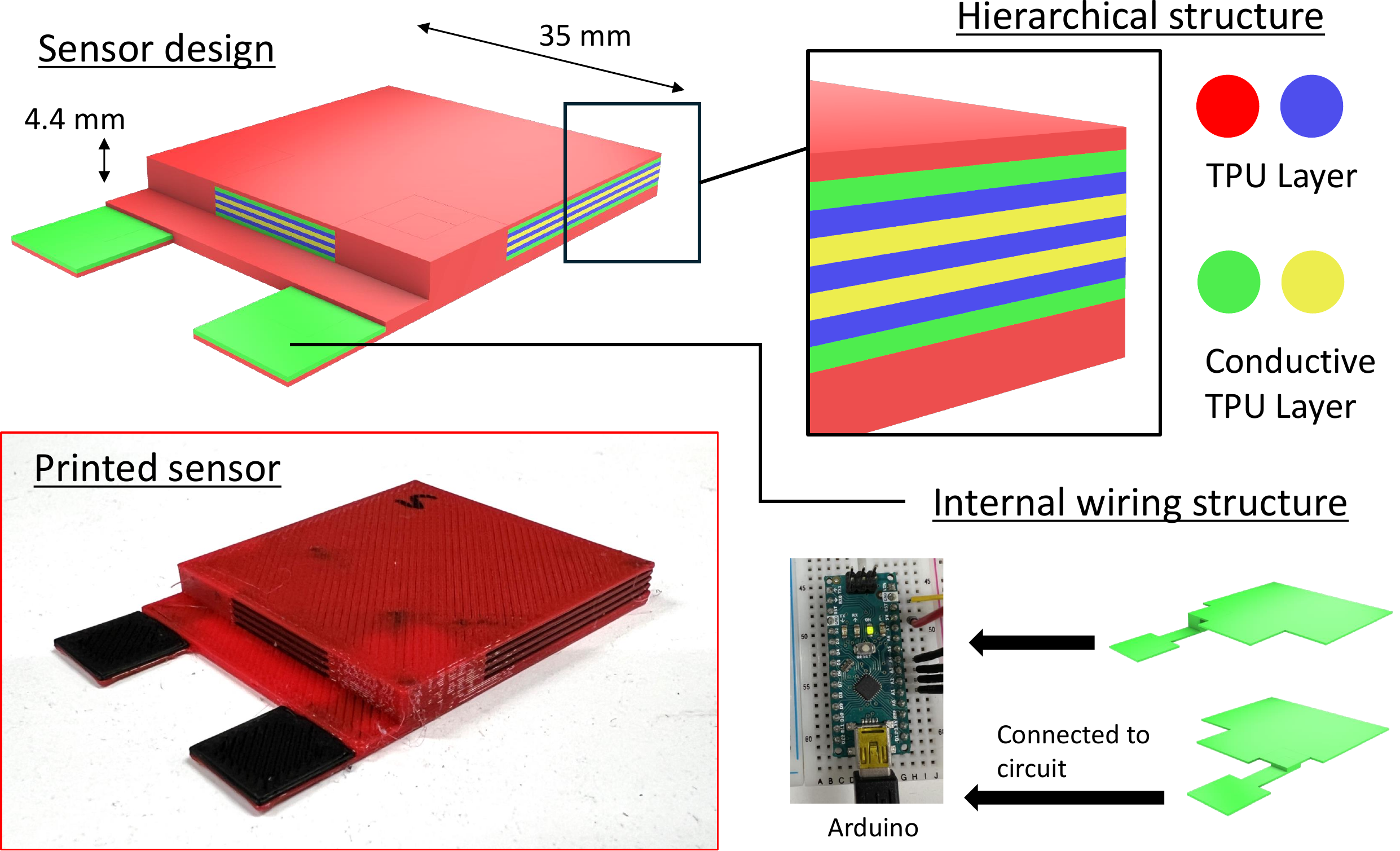} 
  \caption{ 
    The structure of M3D-skin.
    The upper part of the sensor performs sensing through a layered structure composed of conductive and non-conductive
    filaments.
    The lower part of the sensor includes wiring layers, which connect to the circuit. 
  } \label{sensor_structure}
\end{figure}

\subsubsection{Sensor Layer}
The structure of the sensor layer is shown in \figref{sensor_layer}.
At the core of this layer, a sensing structure is created by alternately stacking conductive and non-conductive
filaments with a specific number of layers and infill pattern (\figref{sensor_layer}[A]).
In the 3D design data, the conductive filament layers are completely separated.
Unlike standard 3D printing settings, this layer does not include solid infill on the top or bottom surfaces.
Therefore, at the transition points between conductive and non-conductive filaments, the sparse infill patterns come
into direct contact.
Additionally, solid infill conductive filament layers are placed above and below this hierarchical structure to
encapsulate it (\figref{sensor_layer}[B]).
The resistance measurement during compression is performed in the section between these upper and lower solid infill
layers.

\begin{figure}[t]
  \centering
  \includegraphics[width=1.0\columnwidth]{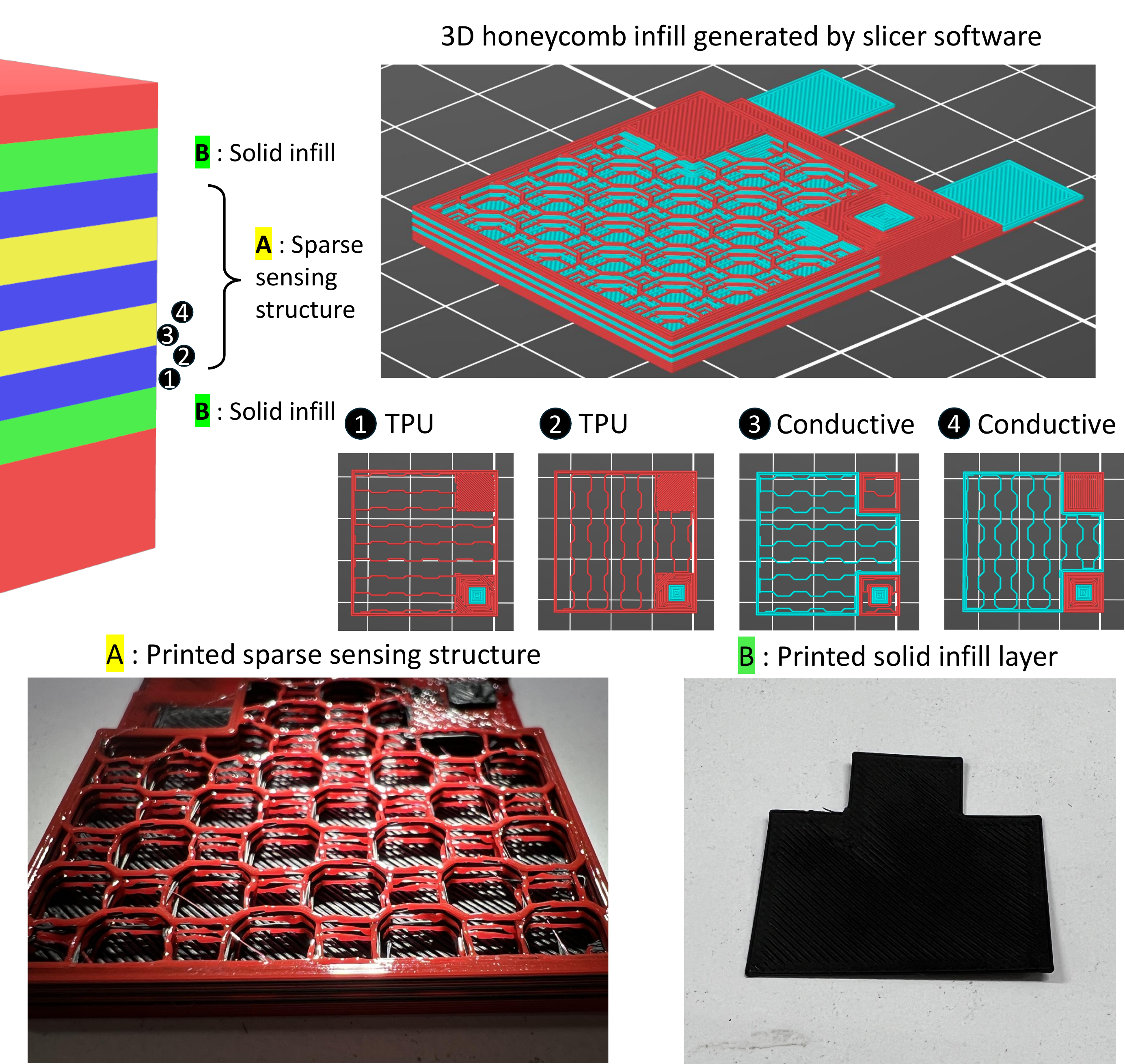}
  \caption{
    The structure
    of the sensor layer: (A) The sparse infill sensing layer, and (B) The solid infill layers above and below the
    sensing layer.
    The sparse pattern structure utilizing infill in (A) is illustrated in the top-right of the figure, showing both the
    overall shape and the shape of each layer.
    The lower section presents the printed results of (A) and (B).
  }
  \vspace{-3mm}
  \label{sensor_layer}
\end{figure}

\subsubsection{Wiring Layer}
The wiring layer is shown in \figref{wiring_layer}.
This structure connects the upper and lower solid infill layers of the sensor layer to the underlying wiring, enabling
connection to the circuit.
This allows terminal connections to the circuit can be placed at any desired location,
regardless of the overall shape of the sensor.

\begin{figure}[t] 
  \centering 
  \includegraphics[width=1.0\columnwidth]{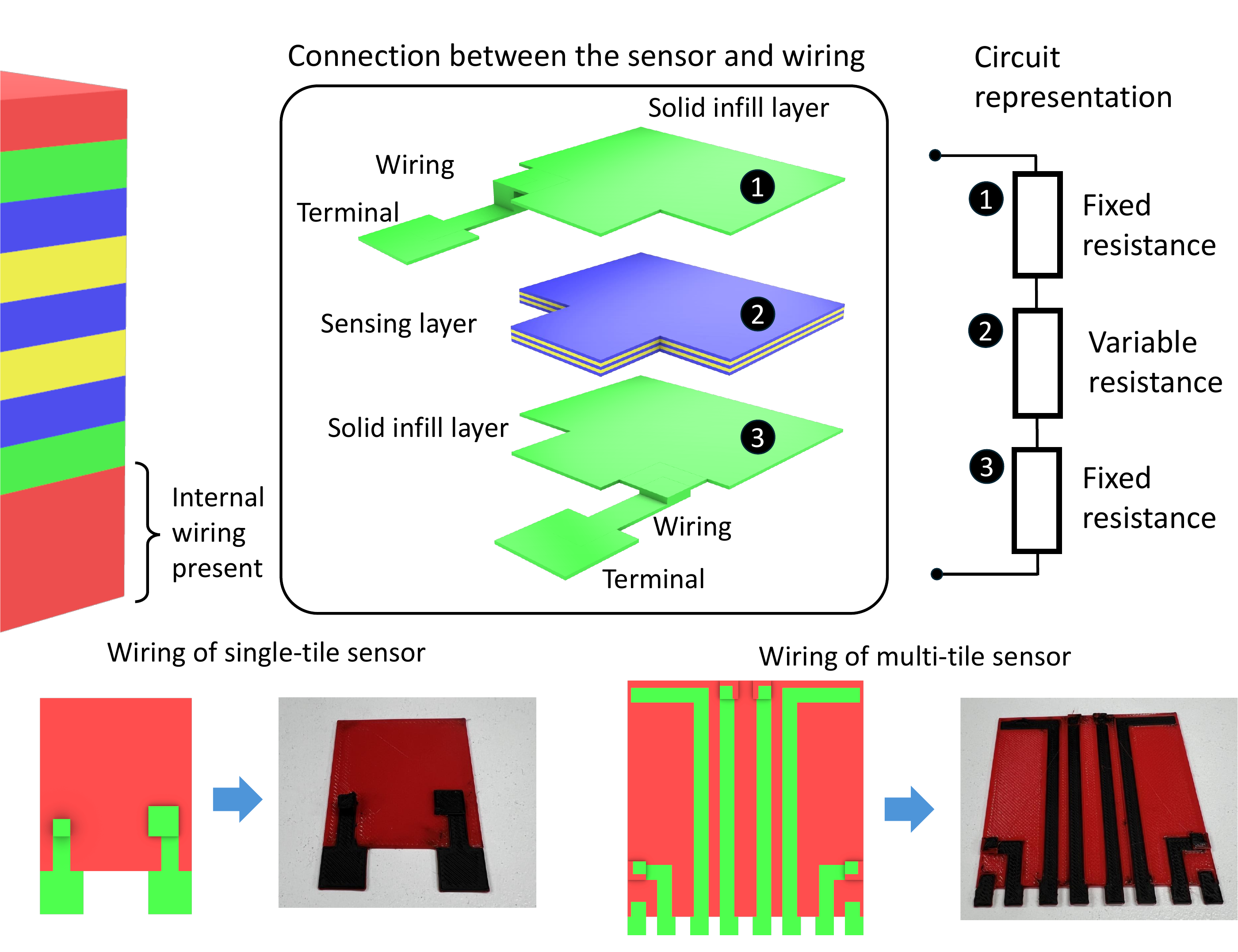} 
  \caption{ 
    The structure
    of the wiring layer.
    It connects the upper and lower infill layers of the sensor layer to the terminals.
    The resistance between the terminals is the sum of the resistance of the infill layers, wiring, and sensor layer.
    The lower section of the figure shows wiring examples for a single-tile sensor and a four-tile sensor. 
  } \label{wiring_layer} 
\end{figure}

\subsubsection{Cover Layer}
The cover layer is placed above and below the sensor and wiring layers to provide surface protection.
This layer is printed using standard TPU filament.

The key design parameters are listed in \tabref{design_param}.
\begin{table}[tbhp]
 \begin{center}
  \caption{Baseline Design Parameters}
  \begin{tabular}{cc} \toprule
  Parameter & Value \\ \midrule
  Sensor Layer Conductive Filament Thickness & 0.4 mm \\
  Sensor Layer Non-Conductive Filament Thickness & 0.4 mm \\
  Sensor Layer Number of Patterned Layers & 4 \\
  Wiring Layer Thickness & 0.4 mm \\
  Cover Layer Thickness & 0.4 mm \\
  Layer Height & 0.2 mm \\
  Nozzle Diameter & 0.4 mm \\
  Sensor Layer Infill Density & 10 \% \\
  Sensor Layer Infill Pattern & 3D Honeycomb \\
  Sensor Layer Wall Thickness & 0.8 mm \\
  Infill Density (Outside Sensor Area) & 100 \% \\
  \bottomrule
  \end{tabular}
  \label{design_param}
 \end{center}
 \vspace{-3mm}
\end{table}

}%
{%
このセンサの構造について、全体像を\figref{sensor_structure}に示す。
積層型3Dプリンタを用いて、導電性および非導電性の
TPU (Thermoplastic Polyurethane)フィラメントを用いて印刷される。
このセンサは、圧力を感知するためのセンサ層および、配線をまとめるための配線層、
センサの表面保護のためのカバー層の3つの構造から構成される。

\begin{figure}[t]
  \centering
  \includegraphics[width=1.0\columnwidth]{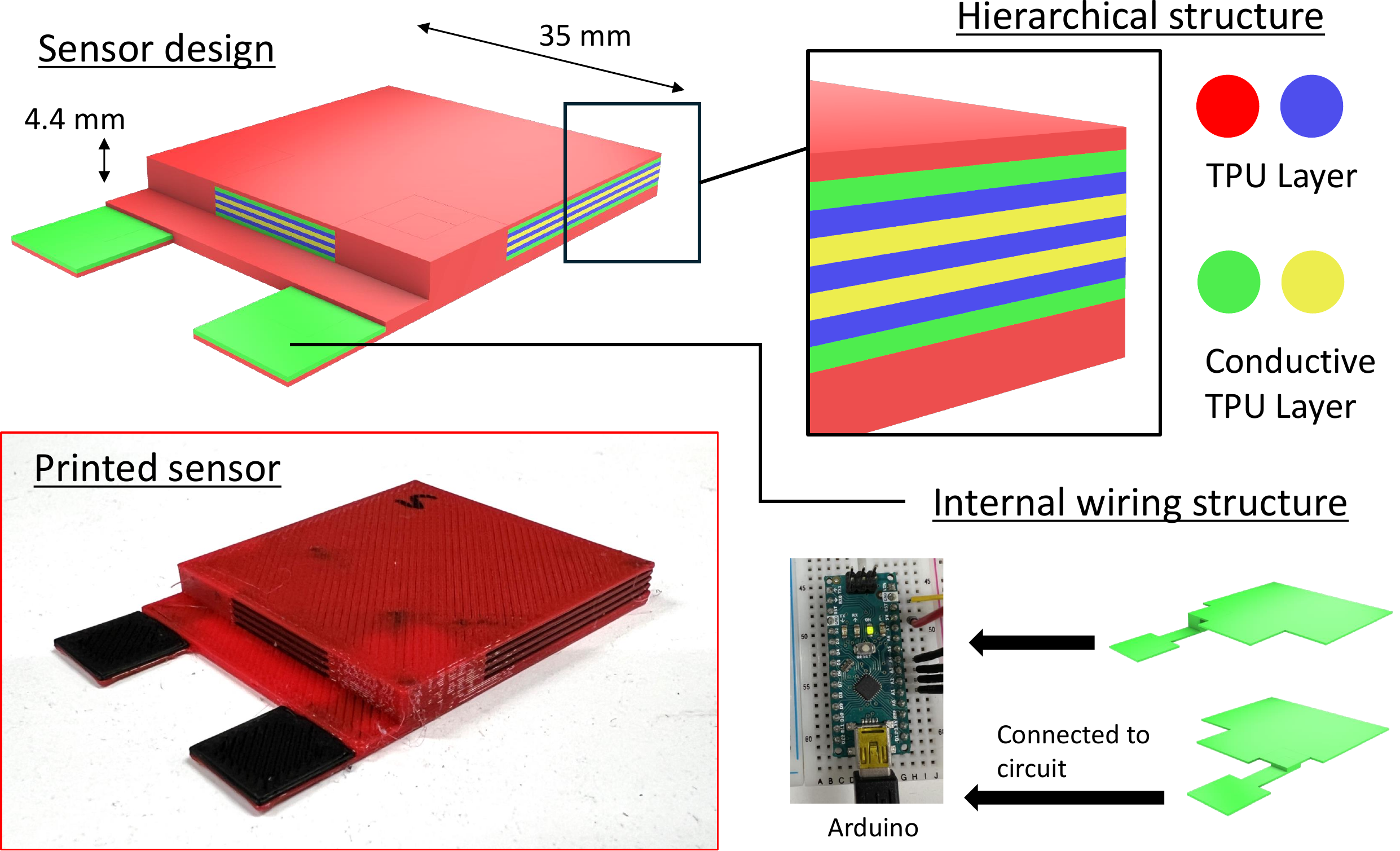}
  \caption{
  センサの構造を示す。
  センサの上部では、導電性・非導電性フィラメントによる階層構造で
  センシングを行う。
  センサの下部には配線のための層があり、
  回路へと接続する。
  }
  \label{sensor_structure}
\end{figure}

\subsubsection{センサ層}
センサ層の構造を\figref{sensor_layer}に示す。
この層の中心には、導電性フィラメントと非導電性フィラメントを、
特定の層数およびインフィルパターンで交互に積層することで作成される、センシングのための
層が存在する(\figref{sensor_layer}[A])。
設計時の3Dデータにおいては、導電性フィラメント層同士は完全に分離している。
なお、これらの層では、通常の3Dプリント設定とは異なり、底面および上面のソリッドインフィルは
設けない。ゆえに、
導電性・非導電性フィラメントが切り替わる部分においては、
疎なインフィルパターン同士が直接接触する。
また、この階層構造を、上下から挟み込むように、ソリッドインフィルの導電性フィラメント
の層を設ける(\figref{sensor_layer}[B])。
圧縮時に抵抗値を計測するのは、この上下のソリッドインフィル層の間の部分となる。

\begin{figure}[t]
  \centering
  \includegraphics[width=1.0\columnwidth]{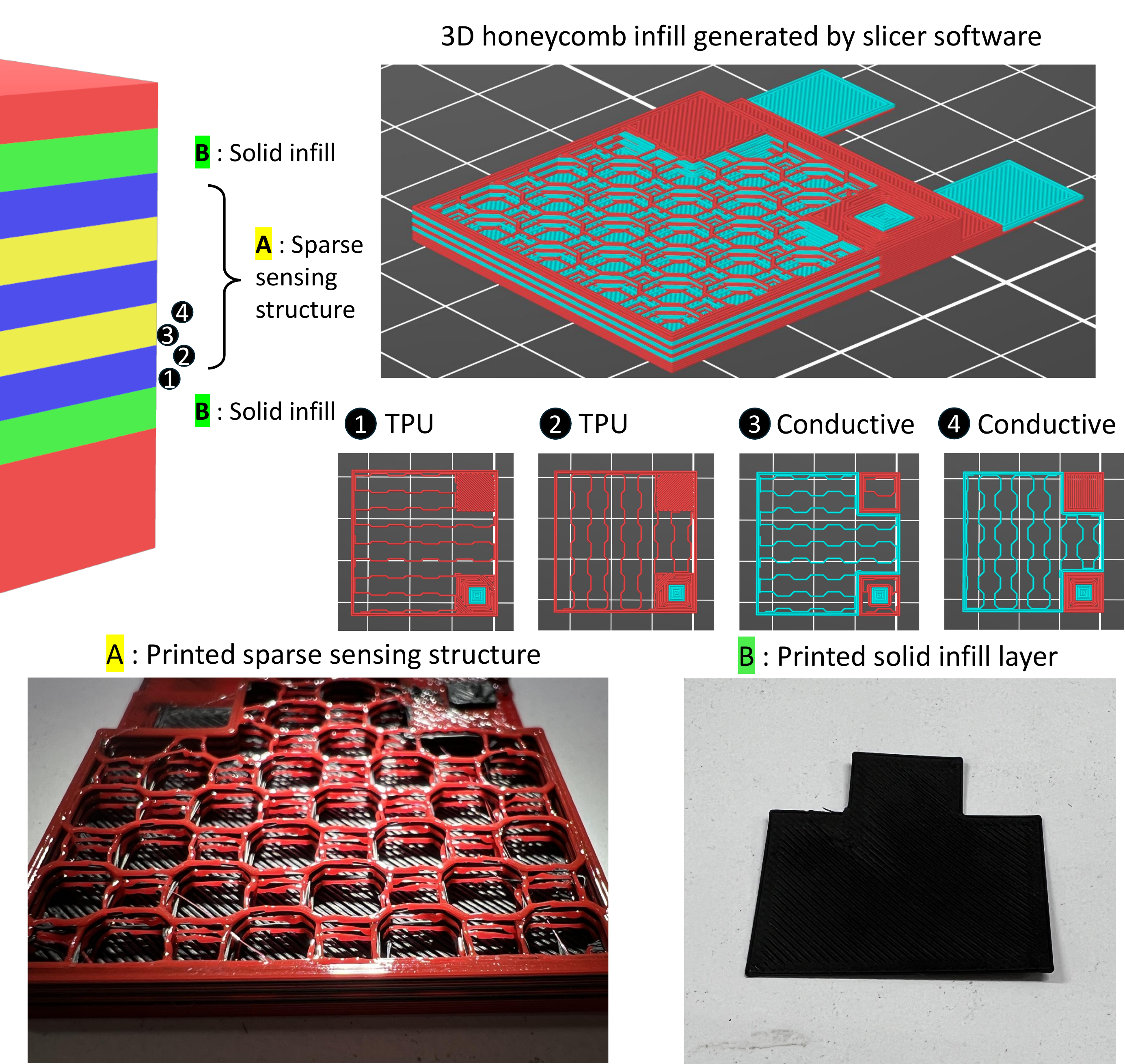}
  \caption{
  センサ層の構造について、(A) センサ層の構造、(B) センサ層の上下のソリッドインフィル層
  を示す。特に、Aのインフィルを活用した疎なパターン構造は、図の上部右に全体および1層ごとの形状を
  示す。
  図の下側に、A,Bを印刷した結果を示す。
  }
  \label{sensor_layer}
\end{figure}

\subsubsection{配線層}
配線層の構造を\figref{wiring_layer}に示す。
この構造は、
センサ層における上下のソリッドインフィル層それぞれから、下層へと配線を繋げて、
回路へと接続するための構造である。
この配線層が存在することで、
センサの全体形状にかかわらず、
回路と接続するための端子を任意の場所に設けることが可能となる。

\begin{figure}[t]
  \centering
  \includegraphics[width=1.0\columnwidth]{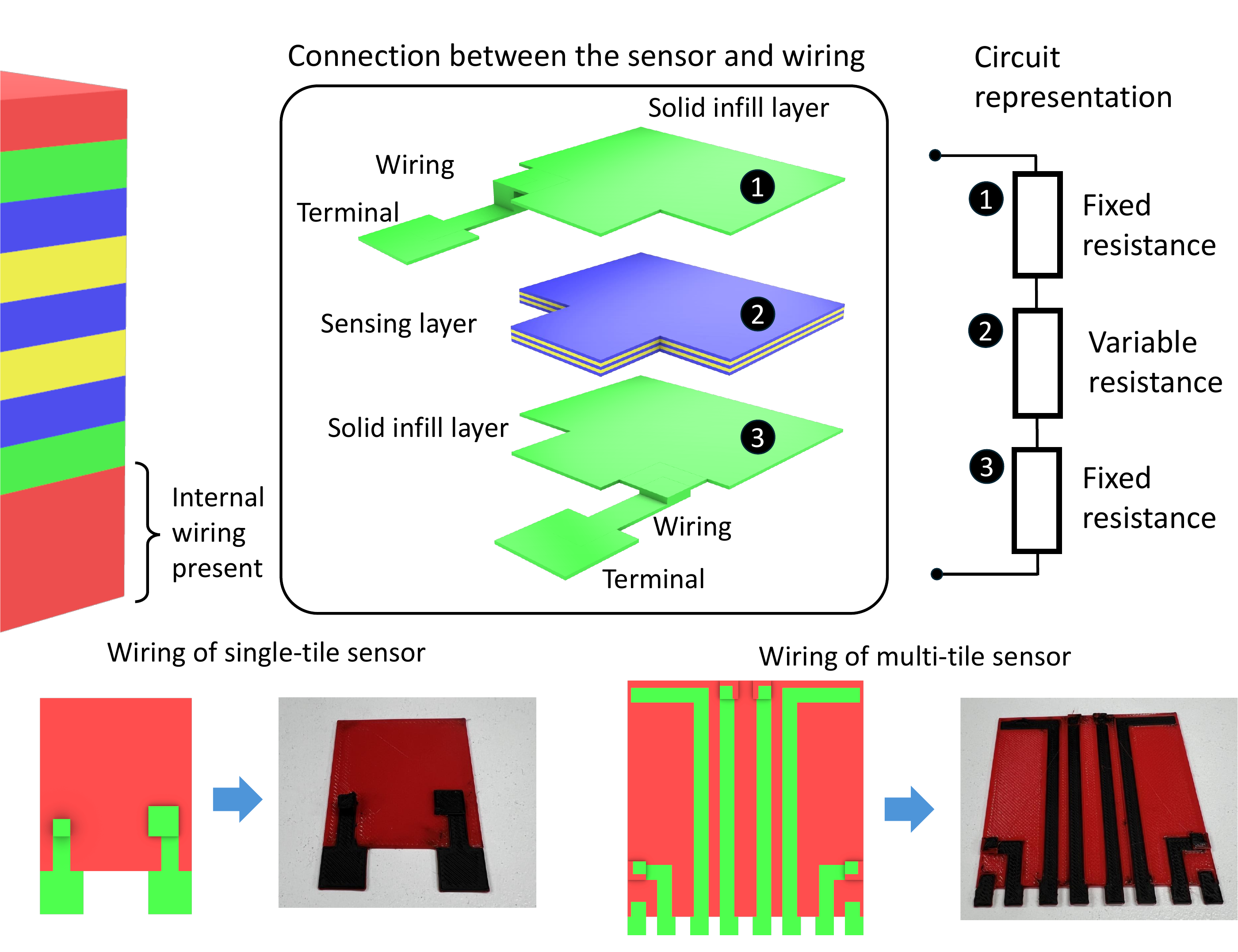}
  \caption{
    配線層の構造を示す。
    センサ層上下のインフィルレイヤを、端子へと繋ぐ役割を有する。
    端子間の抵抗は、インフィルレイヤや配線の抵抗値と、センサ層の抵抗値の和となる。
    図の下側に、1タイルセンサの配線例、4タイルセンサの配線例を示す。
  }
  \label{wiring_layer}
\end{figure}

\subsubsection{カバー層}
センサ層、配線層の上下には、表面の保護のためのカバー層を設ける。
この層は、通常のTPUを用いて印刷される。

このセンサにおいて、基準となる設計パラメータを\tabref{design_param}に示す。
% TODO パラメータと変数名を、表と図で一致させたい

\begin{table}[tbhp]
 \begin{center}
  \caption{基準の設計パラメータ}
  \begin{tabular}{cc} \toprule
  Parameter & Value \\ \midrule
  センサ層 導電性フィラメント部厚み & 0.4 mm \\
  センサ層 非導電性フィラメント部厚み & 0.4 mm \\
  センサ層 階層のパターン数 & 4 \\
  配線層 厚み & 0.4 mm \\
  カバー層 厚み & 0.4 mm \\
  積層の厚み & 0.2 mm \\
  ノズル径 & 0.4 mm \\
  センサ層インフィル密度 & 10 \% \\
  センサ層インフィルパターン & 3Dハニカム \\
  センサ層ウォール & 0.8 mm \\
  センサ部分以外のインフィル密度 & 100 \% \\
  \bottomrule
  \end{tabular}
  \label{design_param}
 \end{center}
\end{table}
}

\subsection{Sensor Principle}
\switchlanguage%
{%
An overview of the sensor principle is shown in \figref{sensor_principle}.
In the sensor layer, conductive and non-conductive filaments are alternately stacked.
In the 3D design data, the conductive filament layers are completely separated because they are interleaved with
non-conductive filament layers above and below.
However, when this structure is printed with a sparse infill pattern, overhangs and printing instabilities cause
portions of the conductive filament to protrude into the adjacent non-conductive filament layers.
As a result, even in the undeformed state, the conductive filament layers establish electrical continuity with the
adjacent layers.
When pressure is applied to this structure, the sparse and flexible infill pattern deforms.
This deformation increases the contact area between conductive filament layers, leading to a change in electrical
resistance.

\begin{figure}[t] 
  \centering 
  \includegraphics[width=1.0\columnwidth]{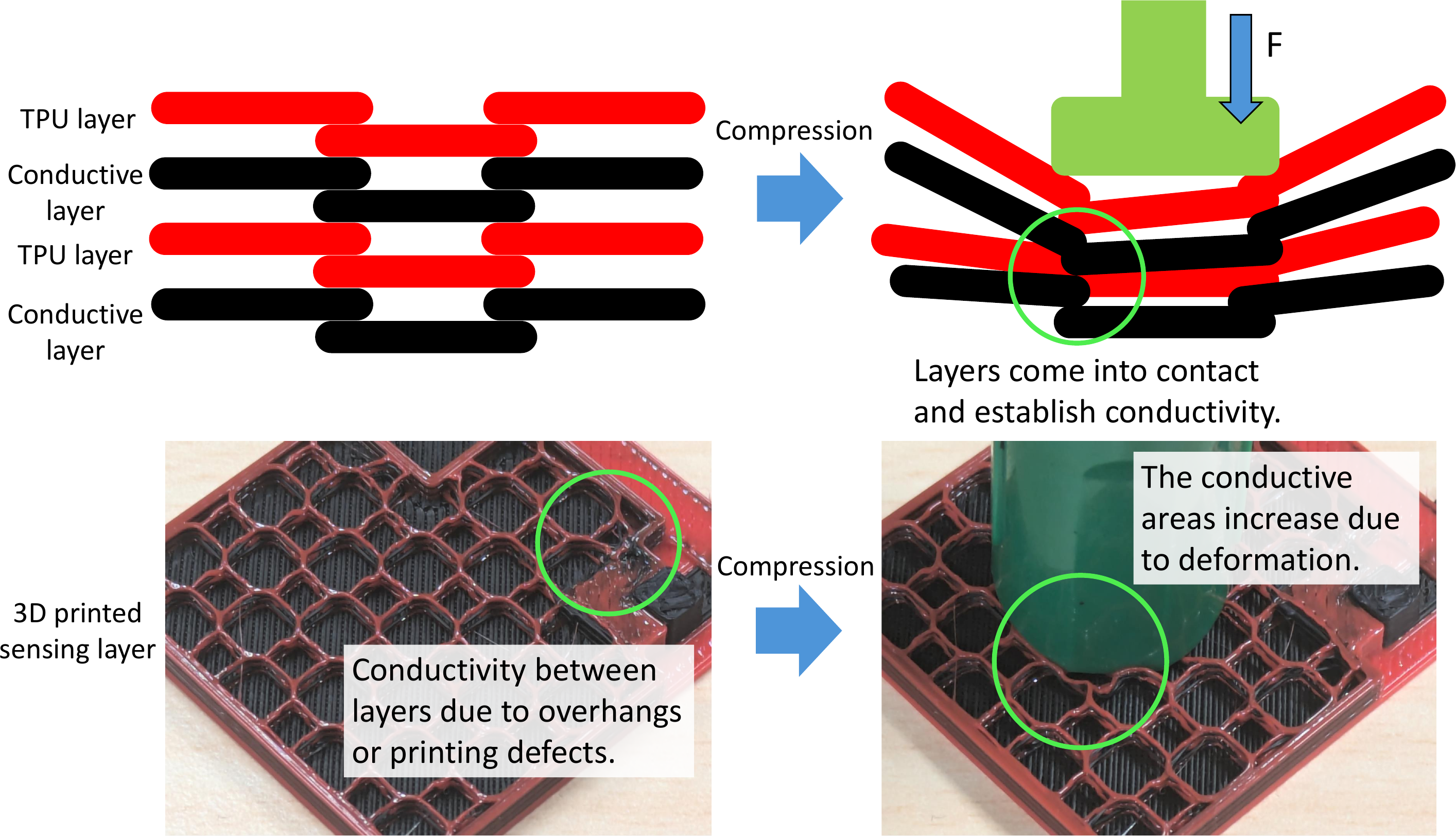} 
  \caption{
  Illustration of the sensor principle, showing how the infill structure deforms under compression. 
  } \label{sensor_principle} 
  \vspace{-2mm}
\end{figure}

It should also be noted that the total resistance measured between the sensor terminals includes both the resistance of the sensor
layer and the resistance of the wiring layer.
Thus, the measured terminal resistance is the sum of the wiring layer resistance and the sensor layer resistance.
A shorter wiring layer reduces its contribution to the total resistance, making the sensor's resistance change more
pronounced and improving its usability.
}
{%
センサの原理について、概要を\figref{sensor_principle}に示す。
センサ層においては、導電性フィラメントと非導電性フィラメントが交互に積層される。
この構造において、設計時の3Dデータでは、
導電性フィラメントの層は、その上下に非導電性フィラメントの層が存在するために、
導電性フィラメント同士は完全に分離されている。
一方、この構造を、インフィルパターンによる疎な構造として印刷すると、
インフィルパターンの内部にはオーバーハングや印刷の不安定性によって、
導電性フィラメントの一部が、その上下の非導電性フィラメントの層へとはみ出る。
結果として、変形を受けていない状態においても、導電性フィラメント層は上下の層と
導通した状態となる。
この構造が圧力を受けると、疎で柔軟なインフィルパターン部分が変形する。
この変形により、導電性フィラメント層同士の接触部分はさらに増え、結果として抵抗値が変化する。
\begin{figure}[t]
  \centering
  \includegraphics[width=1.0\columnwidth]{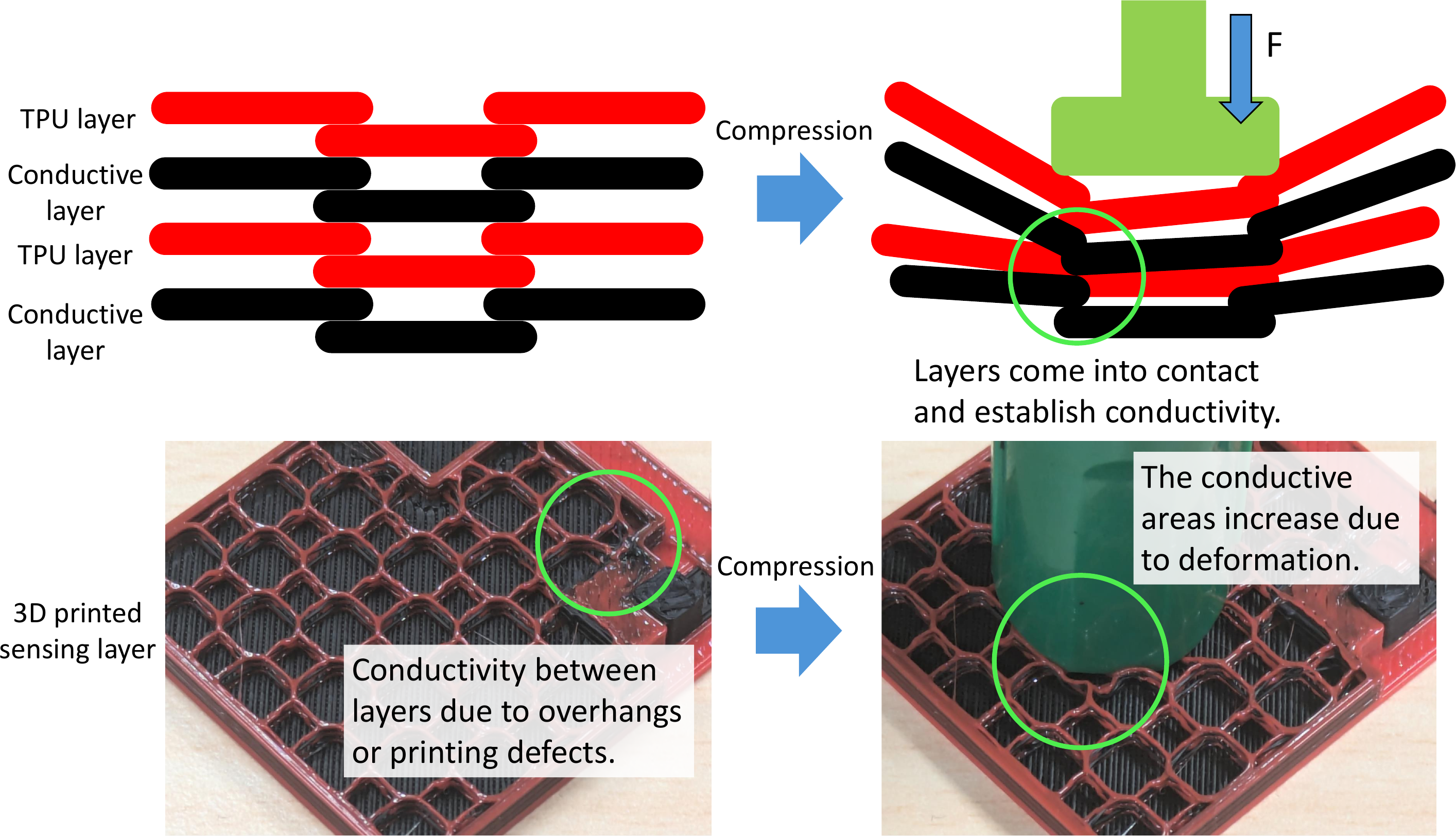}
  \caption{
  センサの原理について、圧縮を受けるとインフィル部分が変形する様子を示す。
  }
  \label{sensor_principle}
\end{figure}

なお、センサの端子間の抵抗として、センサ層だけでなく、配線層にも抵抗が存在する。従って、
端子間の抵抗値として得られる値は、配線層の抵抗値とセンサ層の抵抗値の和となる。
配線層が短いほど、センシングによる抵抗値変化が、端子間の抵抗値に対して大きくなり、
利用に適する。
}

\subsection{Design and Fabrication}
\switchlanguage%
{%
The three-dimensional structure of this sensor is uniquely determined by specifying parameters such as the planar area of its shape and the thickness of the stacked layers.
In this study, to enable rapid design for arbitrary shapes and different parameters, we have systematized the design process using 
parametric modeling with Rhinoceros and Grasshopper.
% \cite{rhino}.
This allows for the fast design of sensor layer structures tailored to specific applications.

For 3D printing, we used the Prusa XL 5-toolhead\cite{prusa},
a multi-material, tool-changer-based 3D printer.
The non-conductive filament used for printing was TPU 95A, while the conductive filament (Conductive Filaflex) was TPU 92A.
The sensor is fabricated in a single integrated printing process, eliminating the need for preprocessing, post-processing, or assembly.
}
{%
このセンサは、センサの形状に関する、平面上の領域および、
積層の厚みなどのパラメータを定めることで、
その三次元的な構造が一意に決定される。
本研究においては、任意の形状や異なるパラメータに対しても
高速に設計を実現するために、
Rhinoceros\cite{rhino}を用いた
パラメトリックモデリングによる設計のシステム化を行っている。
これにより、アプリケーションに合わせたセンサ階層構造を、高速に設計することが可能となっている。

3Dプリンタとして、ツールチェンジャー方式のマルチマテリアル積層3Dプリンタである、
Prusa XL 5-toolhead\cite{prusa}を用いた。
印刷に使用した非導電性フィラメントはTPU95A, 
導電性フィラメントはTPU92Aである。
センサの制作に関して、一体印刷という1プロセスで完成し、
前処理や後処理、組み立てが不要である。
}

\section{Experiment}\label{sec:experiment}
\subsection{Sensor Characterization}
\switchlanguage%
{%
First, we fabricated M3D-skin based on the shape shown in \figref{sensor_structure} 
and the parameters listed in \tabref{design_param}, and measured its characteristics.
A force was applied to the center of the sensor using a force gauge with a 15 mm-diameter circular tip, 
and the changes in resistance were measured via a voltage divider circuit connected to an Arduino.
The results are presented in \figref{s_plot}.
\begin{figure}[t]
  \centering
  \includegraphics[width=1.0\columnwidth]{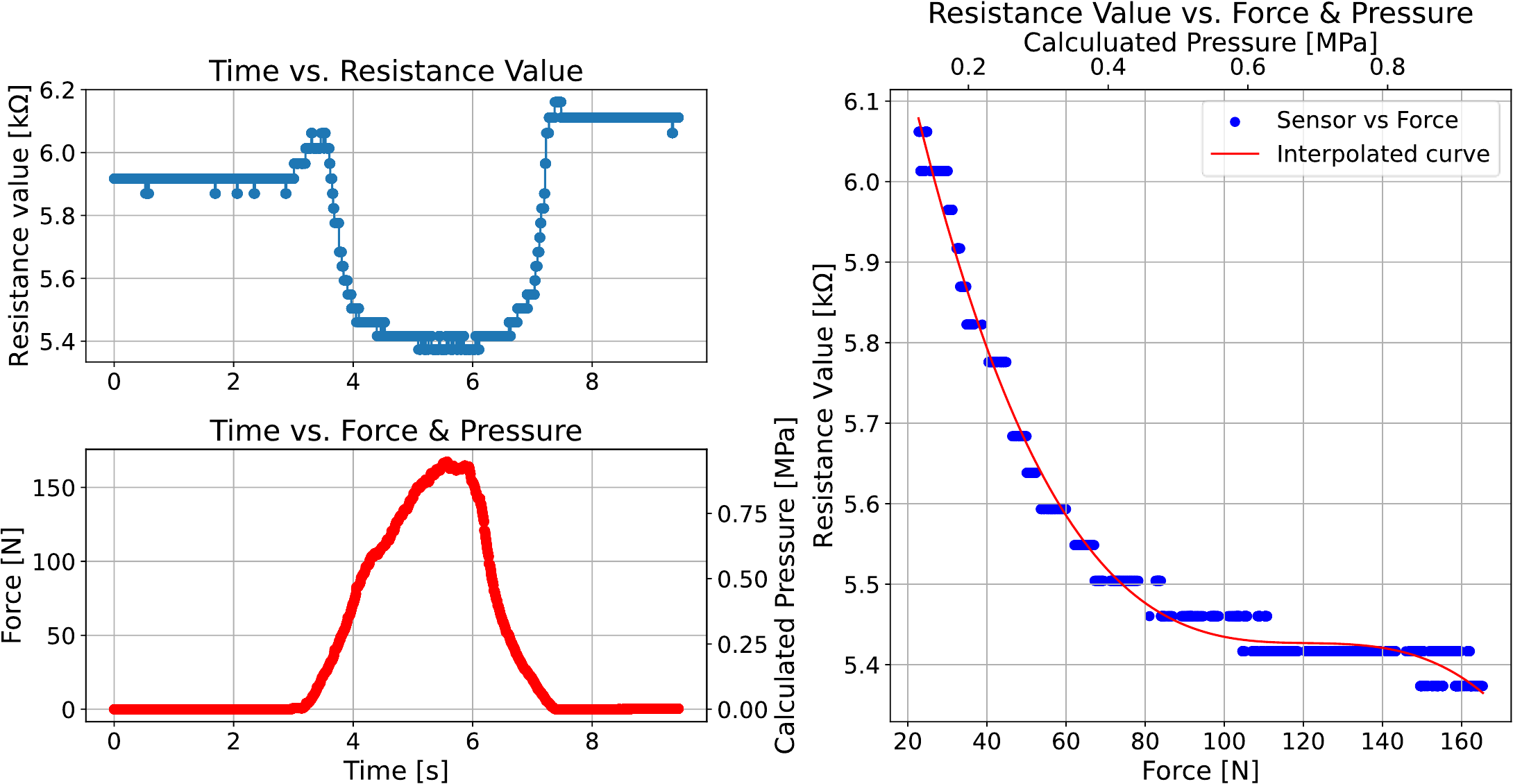}
  \caption{
  Change in resistance when the sensor is subjected to an external force.
  }
  \vspace{-3mm}
  \label{s_plot}
\end{figure}

On the left side of the figure, the time variations of force and resistance are shown.
In the undeformed state, the resistance is about 5.9 k $\Omega$.
As force is applied, the resistance increases to around 6.1 k $\Omega$
up to approximately 25 N, then decreases to about 5.4 k $\Omega$
up to 160 N.
When the applied force is reduced, the resistance returns, reaching 6.1 k $\Omega$
once the force is back to 0 N.
Even when no external force is applied, the resistance remains roughly 0.2 k $\Omega$
higher than the initial state, but gradually decreases over time until it returns to its original value.
This residual resistance change is believed to occur because internal deformation 
remains after a large force is removed and only gradually reverts to its original state.

Next, the right side of \figref{s_plot} shows the relationship between force (and pressure) and resistance in the interval from 
t = 3.5 s to t = 5.5 s, during which the force is applied.
Although the resistance change is significant up to about 100 N and 0.6 MPa, 
it becomes smaller thereafter, gradually changing up to around 160 N.
Compared to the initial resistance of around 6 k $\Omega$,
the change is about 0.6 k $\Omega$,
which is sufficiently large for the Arduino and voltage divider circuit to measure and detect.

\subsection{Relationship Between Structure and Characteristics}
The characteristics of the sensor with different structural configurations are shown in \figref{3_plot} and \figref{pattern}.

\subsubsection{Structural and Parameter Modifications}
First, the left side of \figref{3_plot} shows the characteristics when the layered structure is removed, and the entire sensor layer is made of conductive filament.
In this case, the resistance change is extremely small, demonstrating that the layered structure of conductive and non-conductive filaments is essential for the sensor to function properly.
Next, the center of \figref{3_plot} presents the case where the number of layered structures is increased.
The baseline parameter consists of four conductive layers, but when the number of layers is reduced to three, the resistance change becomes smaller.
Conversely, in the case of five layers, the initial resistance is higher, and the resistance change is particularly large in the 60–100 N range, but the sensor becomes less responsive to smaller forces.
This indicates that increasing the number of layers results in a greater resistance change upon compression, while also modifying the range of detectable forces.
Finally, the right side of \figref{3_plot} shows the characteristics when the sensor layer is printed without walls.
Compared to the baseline parameters with walls, the resistance change is larger, and the sensor is more sensitive to small forces below 20 N.
The absence of walls increases the flexibility of the sensor layer, enhancing sensitivity.
However, due to the reduced rigidity, the printing stability decreases, and black conductive filament contamination is observed on the sensor surface.
While removing the walls improves sensitivity, it also introduces printing instability, which may lead to structural and characteristic inconsistencies.
\begin{figure}[t] 
  \centering 
  \includegraphics[width=1.0\columnwidth]{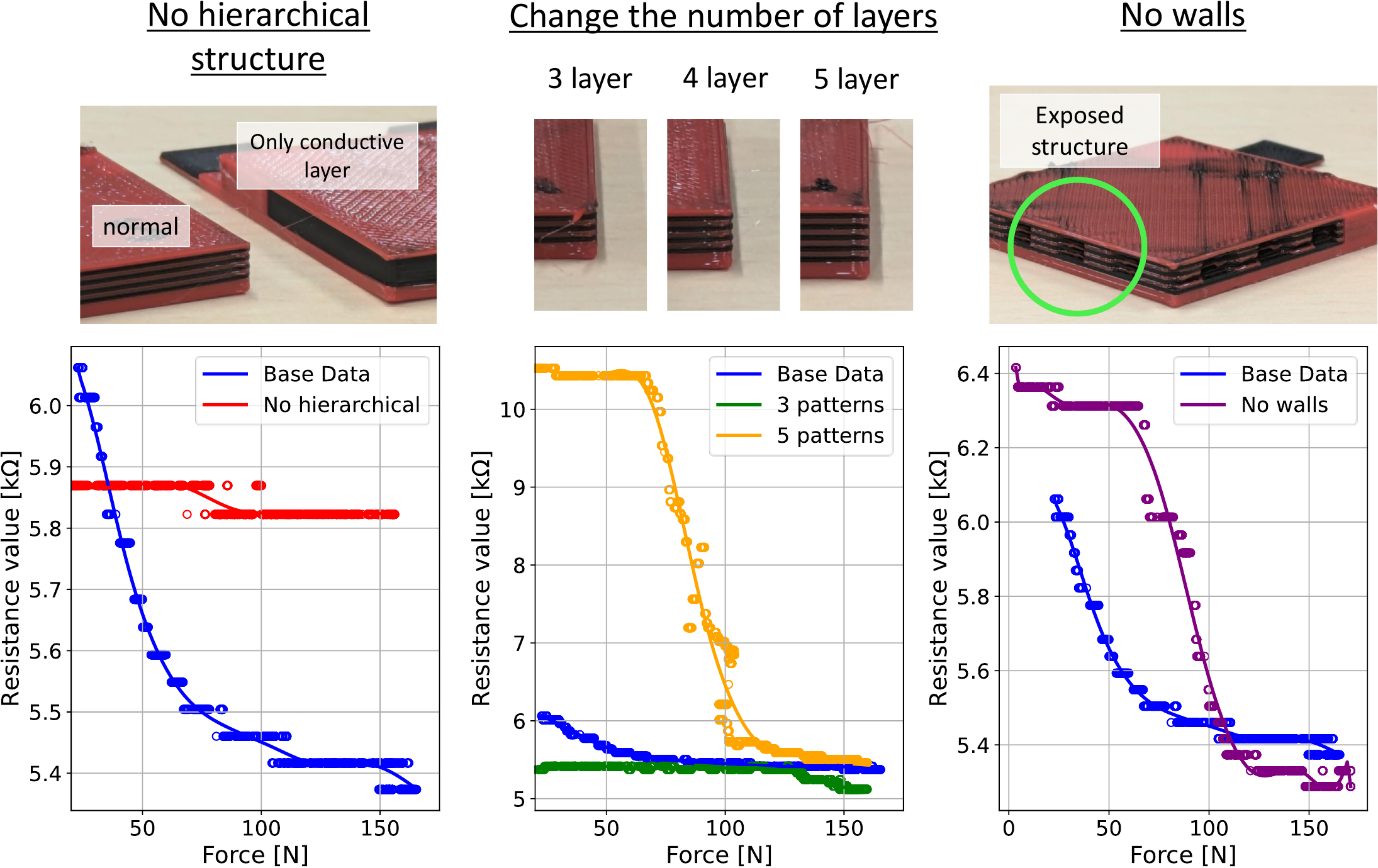} 
  \caption{ 
    Comparison of sensor characteristics with different structures and parameters.
    The figure presents cases where the layered structure is removed, the number of layers is modified, and the walls are removed.
    The blue line represents the baseline parameter for comparison. 
  } \label{3_plot} 
  \vspace{-1mm}
\end{figure}

\subsubsection{Effect of Infill Parameter Changes}
The changes in characteristics when modifying the infill pattern and density are shown in \figref{pattern}.
\begin{figure}[t] 
  \centering 
  \includegraphics[width=1.0\columnwidth]{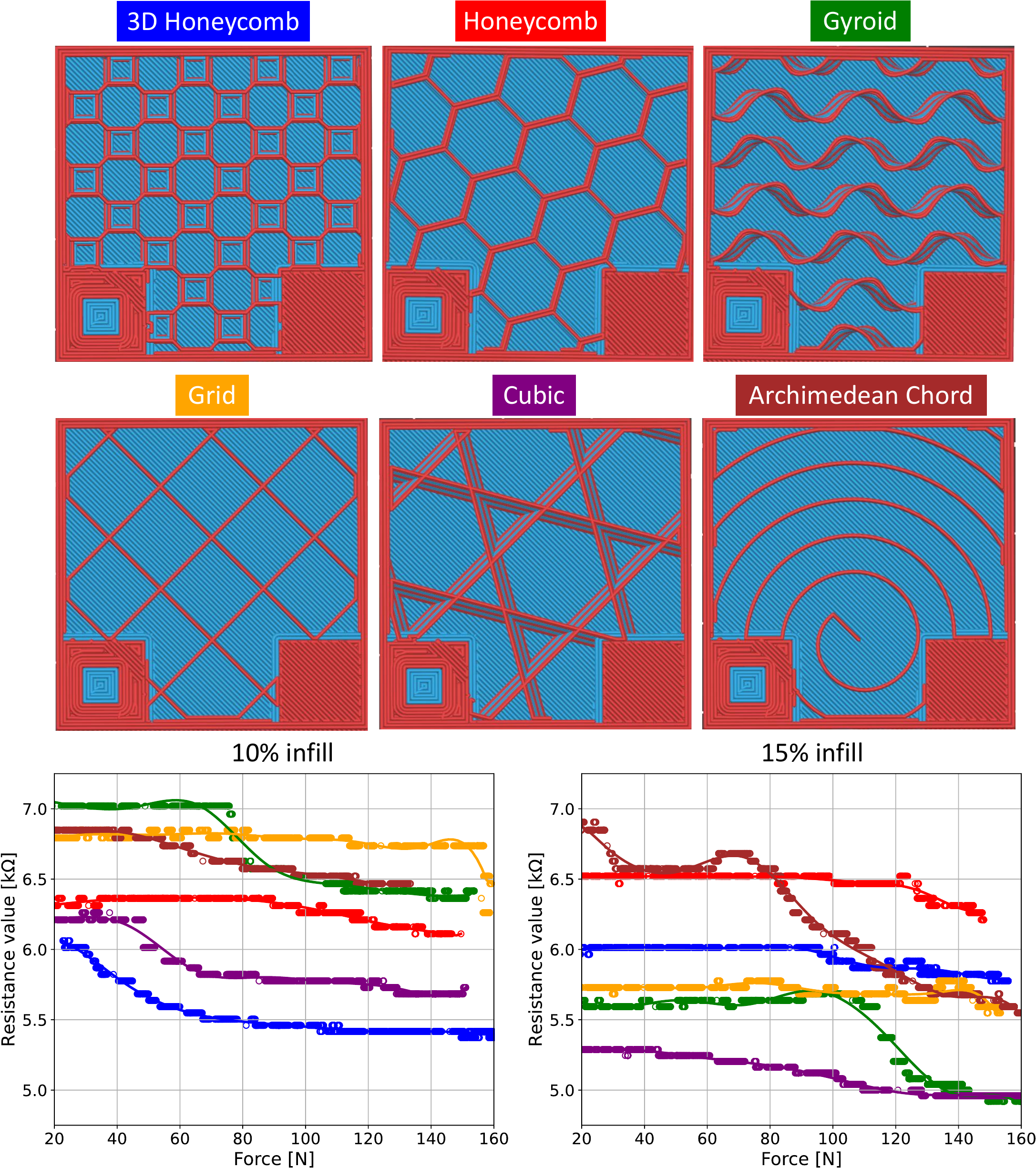} 
  \caption{ 
    Comparison of characteristics when varying infill patterns and densities. 
    The colors of the infill names at the top correspond to the plot colors. 
    Additionally, characteristics for 10\% and 15\% infill densities are shown. 
  } \label{pattern} 
  \vspace{-3mm}
\end{figure}
The sensitivity to force and the magnitude of resistance changes differ depending on the infill pattern.
For 10\% infill density, the 3D honeycomb pattern demonstrates high sensitivity and stable resistance changes up to 100 N, making it well-suited as a sensor.
With 15\% infill density, the sensitivity range of the 3D honeycomb extends from 100 N to 160 N.
The honeycomb pattern exhibits small resistance changes for both densities, likely due to its high structural strength and minimal deformation.
The gyroid pattern shows significant resistance changes, but the range where resistance varies significantly is limited to approximately 40 N.
The cubic pattern, particularly at 15\% infill, achieves a wide sensing range.
The grid and Archimedean chord patterns show resistance variations, but they tend to be unstable, especially at 15\% infill.

These results indicate that selecting an appropriate infill pattern is 
crucial for achieving stable sensor characteristics.
Patterns such as 3D honeycomb, gyroid, and cubic, which change shape across printing layers, 
exhibit greater deformation under compression due to their flexibility, making it easier to obtain resistance changes.
On the other hand, if a sensor that responds only to high pressure is required, 
high-strength infill patterns such as the honeycomb should be chosen.
\subsection{4-Tile Sensor}
To demonstrate the feasibility of using multiple tiles in a sensor, a 4-tile sensor was fabricated, and its resistance change was measured under applied force.
The results are shown in \figref{4tile}, and the wiring layer is illustrated in \figref{wiring_layer}.
In this experiment, the four tiles were sequentially pressed using the tip of a force gauge, and the time variations of resistance and applied force were recorded.
The resistance values of sensors 2 and 3 differ by approximately a factor of two from those of sensors 1 and 4, which is likely due to differences in the resistance of the wiring.
Each sensor exhibited a resistance change of approximately 1 to 2 k$\Omega$ in response to applied force.
These results demonstrate that multiple sensors can be integrally printed and utilized effectively using this approach.

\begin{figure}[t] 
  \centering 
  \includegraphics[width=1.0\columnwidth]{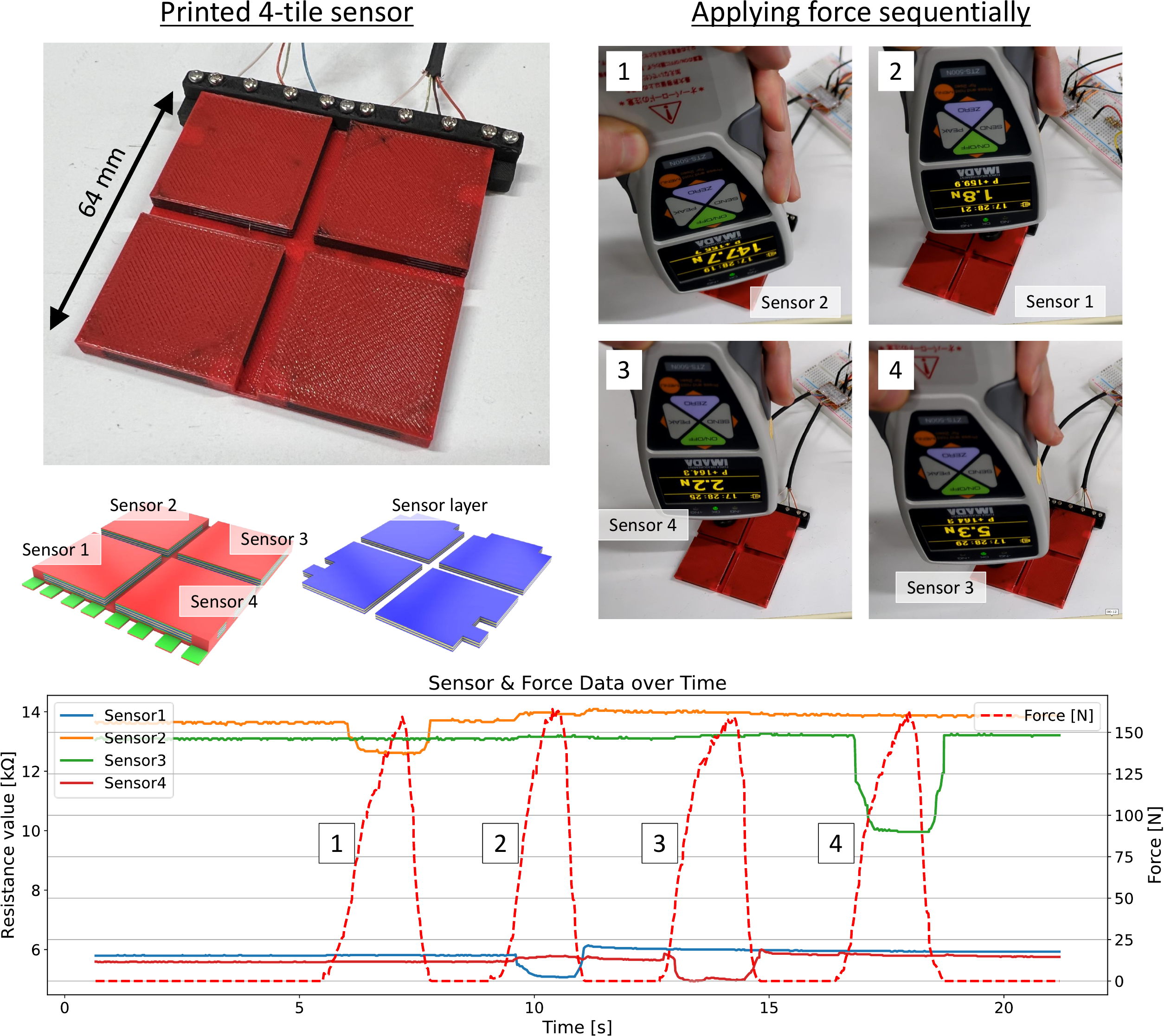} 
  \caption{ Resistance change of the 4-tile sensor under applied force. 
  } \label{4tile} 
  \vspace{-3mm}
\end{figure}

\subsection{Sensor Applications}
% As applications of the sensor, we explored its use as a foot pressure sensor, its integration into a robot hand, and its application in recognition tasks.
As applications of the sensor, we explored its use as a foot pressure sensor and its integration into a robot hand.
\subsubsection{Foot Pressure Sensor}
The application of the sensor as a foot pressure sensor is shown in \figref{sole}.
In this experiment, a six-tile sensor was designed to match the shape of the sole, attached to the foot, and tested during walking and stair climbing.
The pattern variations in sensor readings were measured.
Since the resistance values varied significantly across different tiles, the voltage values of the voltage divider circuit, directly read by an Arduino, were plotted for comparison.
During walking, all sensors exhibited periodic changes in accordance with the walking rhythm.
In contrast, during stair climbing, sensors 5 and 6, located at the heel, showed almost no change, while the other four sensors exhibited regular variations.
These results indicate that the sensor can capture sufficient pattern variations corresponding to human movements.
The proposed sensor is demonstrated to be useful for human motion measurement.
\begin{figure}[t] \centering \includegraphics[width=1.0\columnwidth]{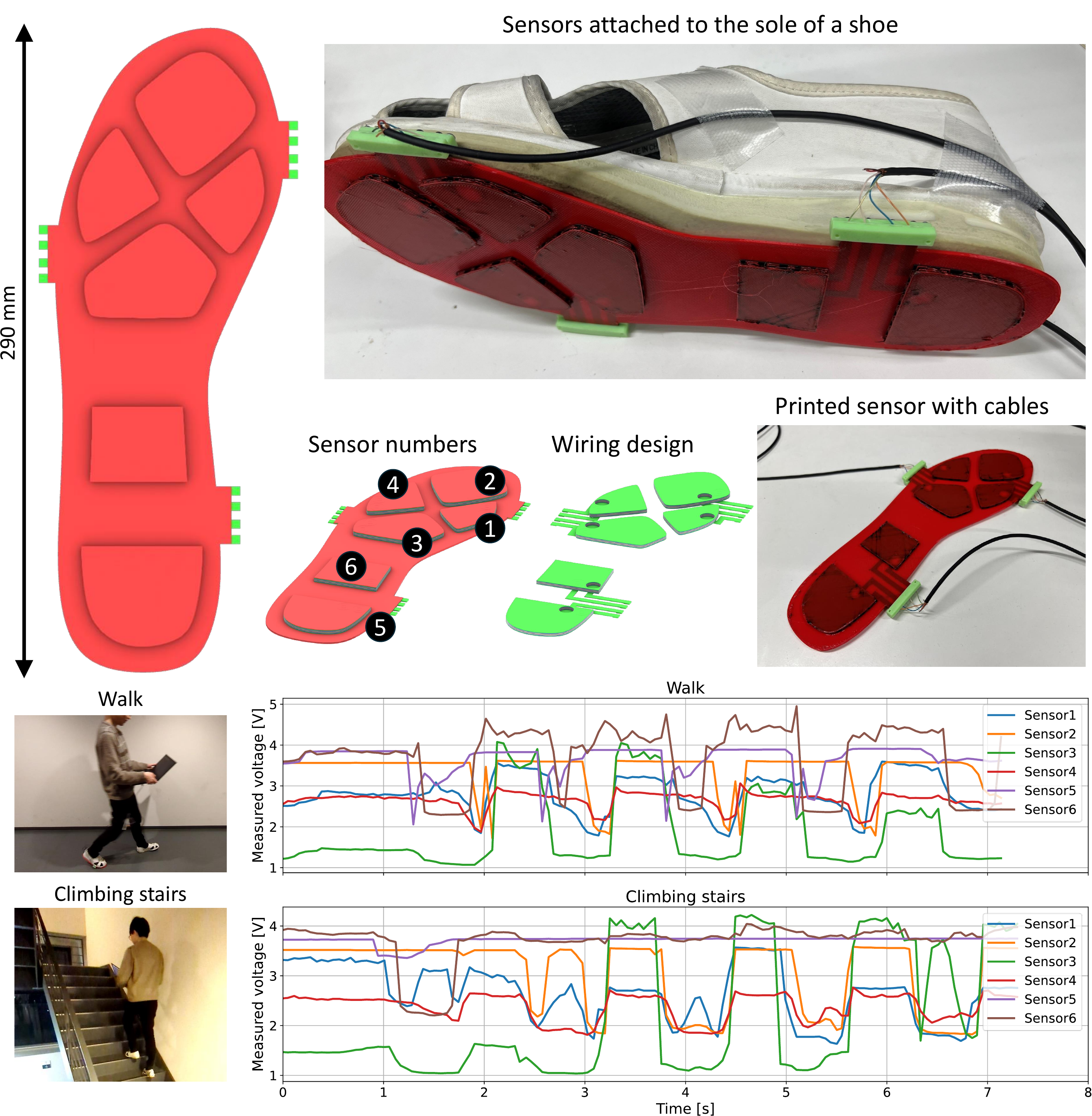} 
  \caption{ 
    Application of the sensor as a foot pressure sensor. 
    The upper part of the figure shows the designed and printed sensor, as well as its attachment to the shoe sole.
    The lower part illustrates the changes in sensor readings during walking and stair climbing. 
  } \label{sole} 
  \vspace{-3mm}
\end{figure}

\subsubsection{Robot Hand}
The use of tactile sensing in a robot hand is shown in \figref{pr2}. In this experiment, a four-tile sensor was 
attached to the right hand of the dual-arm robot PR2 to perform object grasping and recognition tasks. 
The sensor was fabricated using single-step multi-material integrated printing, with PLA filament as the
connection part to the hand, and protrusions were added to the sensor surface to enhance sensitivity.
The middle section of \figref{pr2} illustrates the grasping of four different objects, held either closer 
to the fingertips or deeper in the hand, while the corresponding sensor resistance varied from approximately 1 to 2 k $\Omega$.
The differences in resistance values among the tiles are considered to be due to variations in the resistance of the wiring sections.
Next, the lower section of \figref{pr2} presents an experiment in which the sensor readings during object grasping were used to 
determine whether the object was positioned near the fingertips or deeper in the hand, 
and the robot adjusted its position accordingly to insert the object into a tube. Initially, 
the resistance value of Sensor 2 changed, indicating that the object was located near the fingertips, 
leading to the determination of its position. In the second trial, the resistance value of Sensor 4 changed, 
indicating that the object was positioned deeper in the hand, and the position was adjusted accordingly.
This experiment demonstrated not only the applicability of the sensor for robotic applications but also the 
feasibility of easily integrating sensors through single-step multi-material integrated printing using a 3D printer. 
% Through the four experiments conducted so far, the characterization of the sensor was verified, and 
% it was shown that its properties could be modified by altering its structure and parameters. 
% Additionally, the potential of multi-tile sensors for human body measurements and robotic applications was confirmed, 
% along with the ease of integration through single-step multi-material integrated printing.
\begin{figure}[t] 
  \centering \includegraphics[width=1.0\columnwidth]{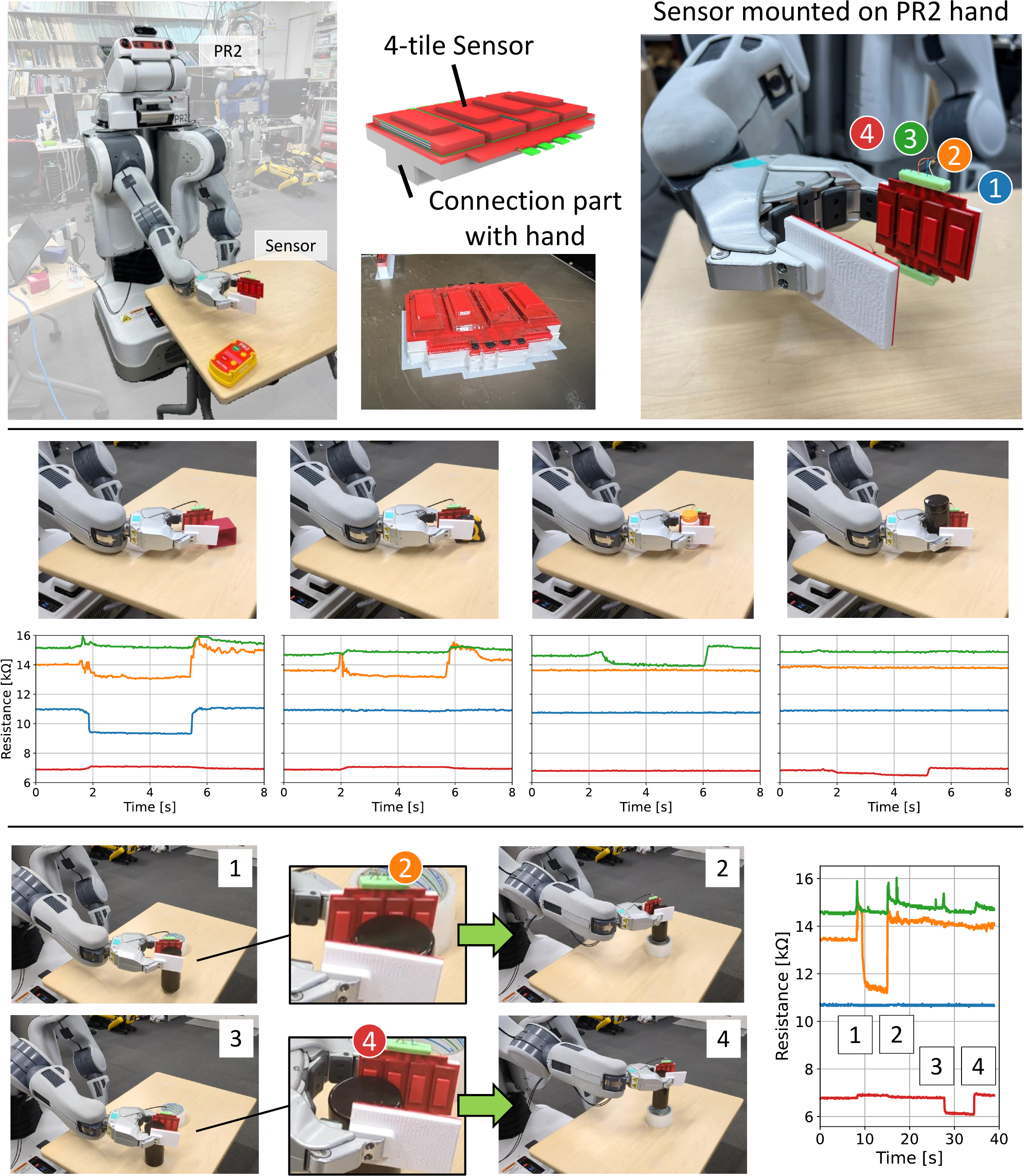} 
  \caption{ Illustration of integration and recognition tasks using a robot hand.
  The top section shows the sensor fabricated through single-step multi-material integrated printing with conductive/non-conductive TPU and PLA, as well as its attachment to the right hand of PR2.
  The middle section displays the grasping of four different objects and the corresponding changes in sensor values.
  The lower section shows the recognition of object positions using the sensor and the subsequent action of inserting the object into a tube.
  } \label{pr2} 
  \vspace{-3mm}
\end{figure}

}%
{%
まず、\figref{sensor_structure}で示した形状および\tabref{design_param}で
示したパラメータに基づいて作成したセンサについて、
その特性を測定した。
センサ中央に、直径15 mm の円形の先端を有するフォースゲージで力を与え、
抵抗値の変化は分圧回路を通してArduinoで測定した。
% TODO : 測定方法
この結果を\figref{s_plot}に示す。
まず、図の左側には、力と抵抗値の時間変化を示す。
変形のない状態では5.9 k$\Omega$であるが、
力が加わると、およそ25N程度までは抵抗値が6.1 k$\Omega$程度まで上昇した後に、
160N程度までの範囲で、抵抗値が5.4 k$\Omega$まで低下する。
力が減少するにつれて抵抗値は戻ってゆき、0 Nとなった時点で6.1 k$\Omega$となる。
外力のない状態でも、初期状態から0.2 k$\Omega$ほど抵抗値が大きくなっているものの、
時間が経つにつれて少しずつ抵抗値が下がり、初期状態に戻る。
この、抵抗値変化の残留は、大きな力を受けた場合に、力がなくなっても内部に変形が
残留し、それが少しずつ戻ることによるものと考えられる。
次に、力を加え始めたt = 3.5 sからt = 5.5 sの間における、力および圧力と抵抗値の
関係を、\figref{s_plot}の右側に示す。
100Nおよび0.6 MPa程度までは抵抗値の変化は大きいものの、それ以降は変化は小さく、
160N程度までの範囲で抵抗値が少しずつ変化する。
初期状態およそ6 k $\Omega$に対して、0.6 k $\Omega$程度の変化が得られており、
Arduinoおよび分圧回路でも十分に変化を計測、感知可能である。

% TODO : 考察
\begin{figure}[t]
  \centering
  \includegraphics[width=1.0\columnwidth]{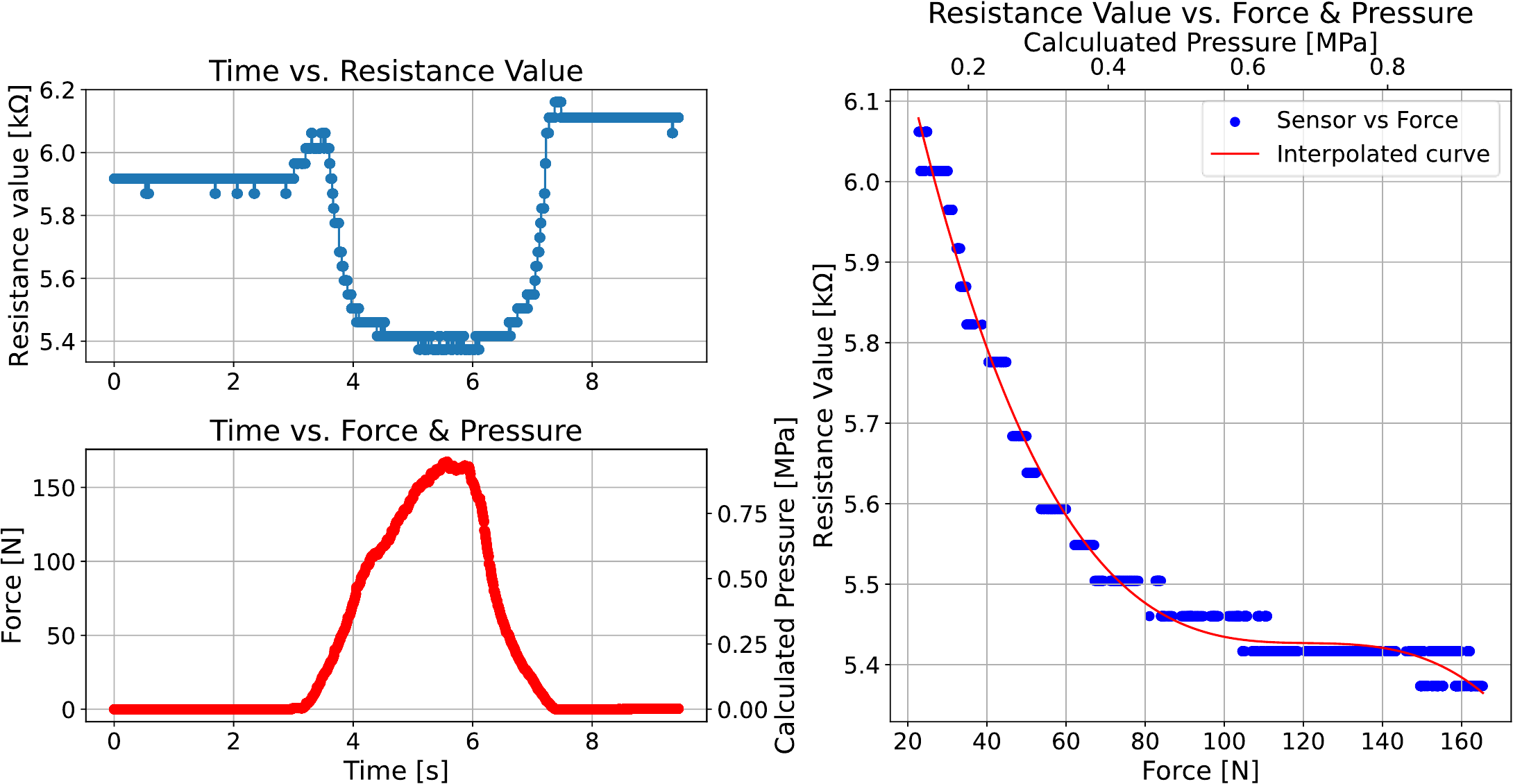}
  \caption{
  センサが外力を受けた際の、抵抗値の変化。
  }
  \label{s_plot}
\end{figure}

\subsection{構造と特性の関係}
センサの構造を変えた場合の特性について、\figref{3_plot}および\figref{pattern}に示す。
\subsubsection{構造・パラメータ変更}
まず、\figref{3_plot}の左側には、階層構造をなくし、
センサ層を全て導電性フィラメントで作成した場合の特性を示す。
この場合、抵抗値変化は非常に小さい。
センサとして利用するために、導電性・非導電性フィラメントの階層構造が
必須であることがわかる。
次に、\figref{3_plot}の中央には、階層構造のレイヤ数を増やした場合を示す。
基準パラメータは4つの導電性レイヤを有するが、
3レイヤの場合は抵抗値変化がより小さくなった。
5レイヤの場合は、通常状態の抵抗値が大きくなっており、
抵抗値変化は60から100Nの部分で特に大きいがそれより小さい力には感知しづらい。
層数を増やすと、圧縮の際の抵抗値変化が大きくなり、同時に
感知可能な力の範囲にも変化が生じることがわかる。
最後に、\figref{3_plot}の右側には、センサ層の印刷時に壁をなくした場合を示す。
壁がある基準パラメータと比べ、抵抗値変化が大きく、また20N以下の小さな力に対しても
敏感に反応する。
壁がないために、センサ層の柔軟性が上がり、感度が高くなっていると考えられる。
一方、剛性が下がるため、印刷の際の安定性が低下し、
センサ表面に黒い導電性フィラメントの汚れがみられる。
壁をなくすことは、センサの感度を向上させる一方で、印刷の不安定性によって
形状や特性が不安定になりうることに留意する必要がある。
\begin{figure}[t]
  \centering
  \includegraphics[width=1.0\columnwidth]{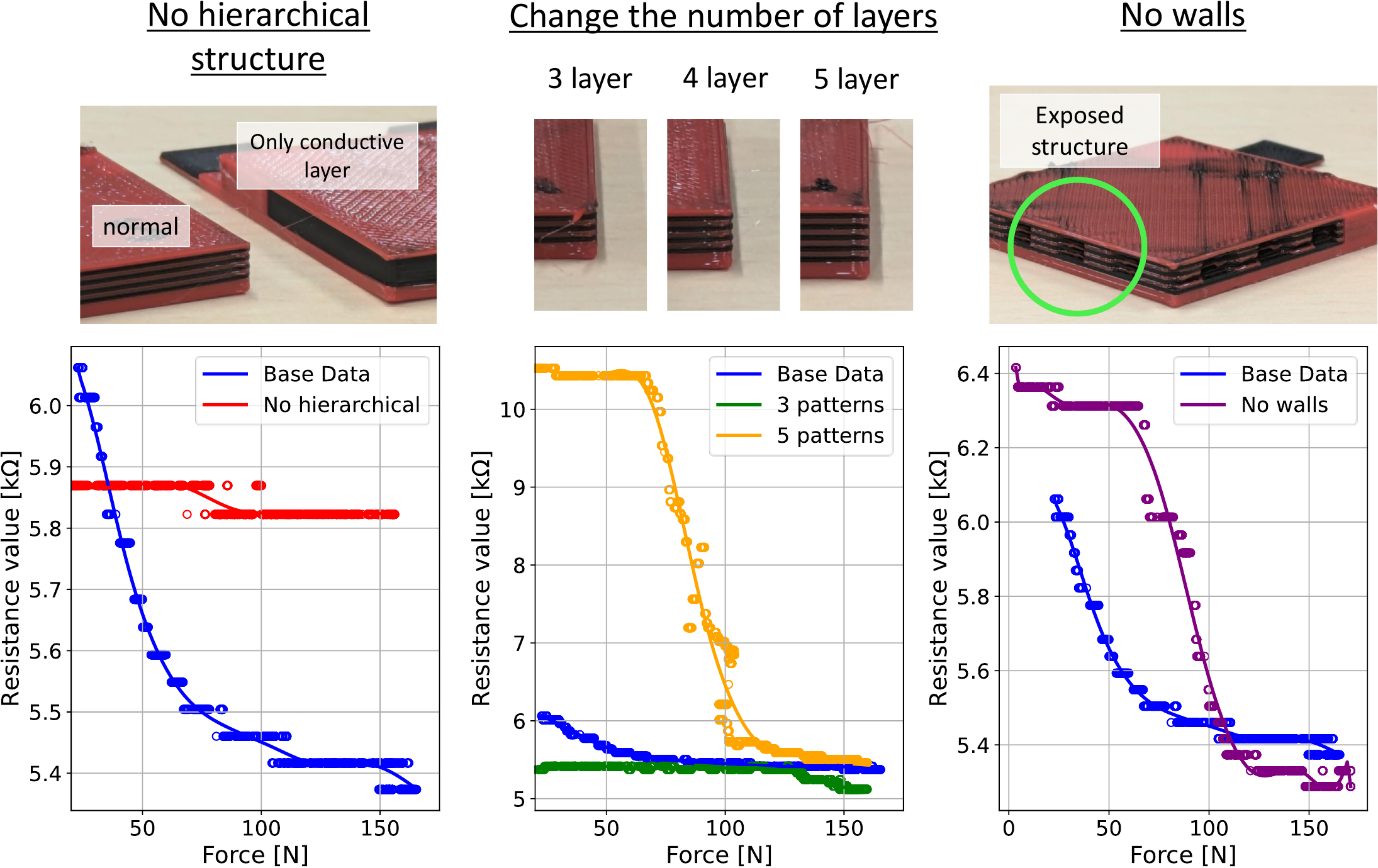}
  \caption{
  構造およびパラメータを変更した場合の特性比較。
  階層構造がない場合、階層数を変えた場合、壁をなくした場合の特性について
  示す。青は比較用の基準パラメータの特性である。
  }
  \label{3_plot}
\end{figure}
\subsubsection{インフィルパラメータ変更}
インフィルパターンおよび密度を変更した場合の特性変化について、
\figref{pattern}に示す。
\begin{figure}[t]
  \centering
  \includegraphics[width=1.0\columnwidth]{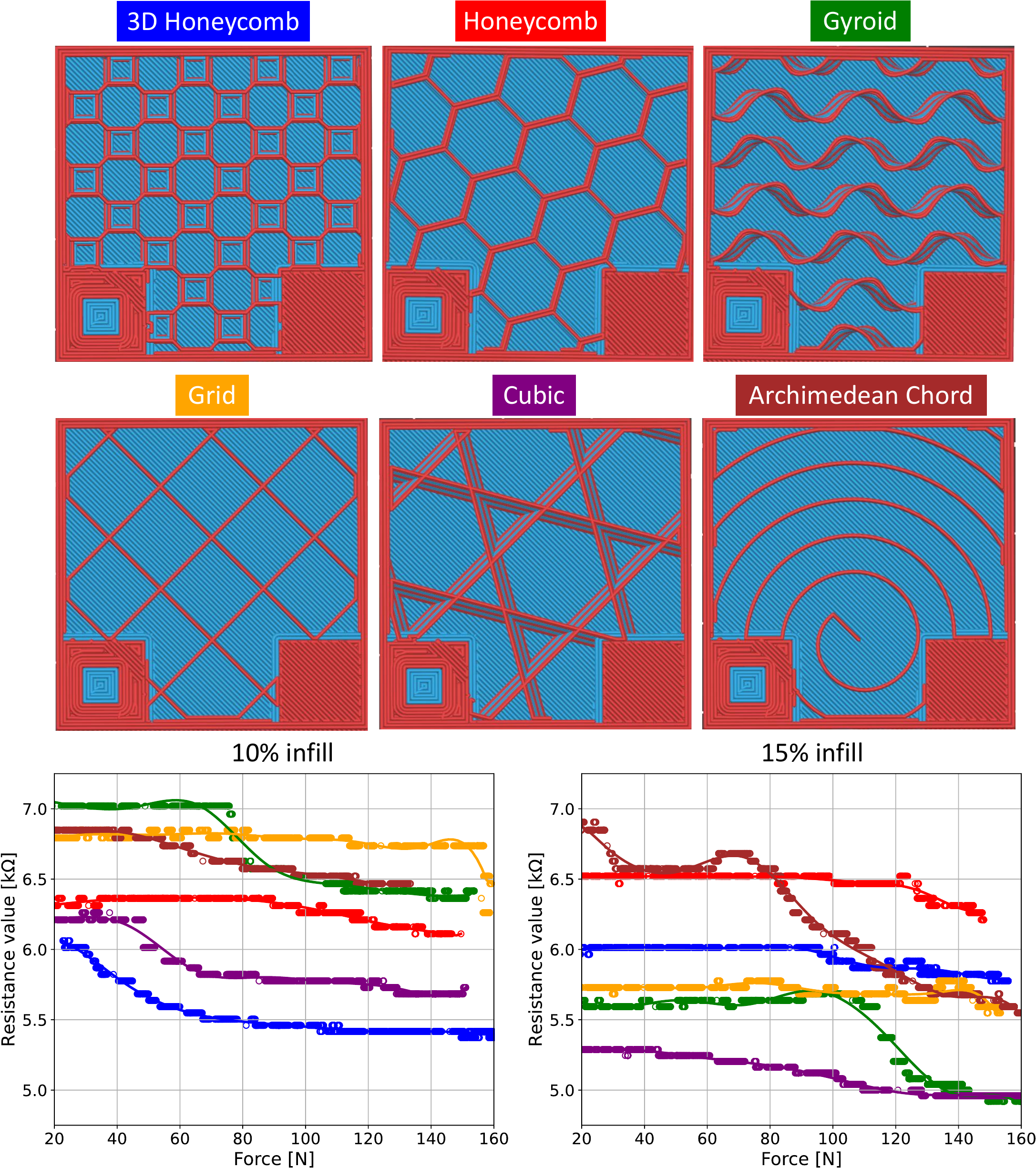}
  \caption{
  インフィルパターンと密度を変えた場合の特性比較。
  上部のインフィル名の色とプロットの色が対応する。
  また、10\%および15\%のインフィルにおける特性を示す。
  }
  \label{pattern}
\end{figure}
インフィルに応じて力を感知しやすい領域および抵抗値変化の大きさが異なる。
3Dハニカムは、10\%のインフィルにおいて、100Nまでの範囲では高い感度と安定した
抵抗値変化を活用でき、センサとしての適性が高い。
15\%のインフィルでは、3Dハニカムの感度幅は100から160Nへと
変化している。
ハニカムパターンはどちらの密度でも抵抗値変化が小さいが、これはパターンの強度が高く
構造の変形が少ないためと考えられる。
Gyroidは抵抗値変化は大きいものの、抵抗値が大きく変化する領域幅が40N程度と低い。
Cubicでは、特に15\%のインフィルにおいて、広い感知幅を得られている。
Grid, Archimedean Chordは、特に15\%において、抵抗値変化はあるものの不安定である。

これらの結果より、センサとして安定した特性を得るためには、
インフィルパターンの選択が重要である。
特に3Dハニカム、Gyroid, Cubicといった、上下の印刷層で形状が変化するパターンは、
その柔軟性から、圧縮時にインフィルの変形が大きく、抵抗値変化を取得しやすいと考えられる。

\subsection{4タイルセンサ}
複数のタイルを有するセンサとしての利用可能性を示すために、
4タイルセンサを作成し、力を与えて抵抗値変化を測定した。
これを\figref{4tile}に示す。
配線層は\figref{wiring_layer}に示した。
この実験においては、4つのタイルを順にフォースゲージの先端で押し込み、
抵抗値および力の時間変化を測定した。
抵抗値はセンサ2,3およびセンサ1,4で2倍ほどの違いがあるが、これは配線部分の
抵抗値の違いによると考えられる。
各センサは力に応じて1から2k $\Omega$程度の抵抗値変化が見られ、
この手法を通して複数センサを一体印刷して利用が可能であることが示された。
\begin{figure}[t]
  \centering
  \includegraphics[width=1.0\columnwidth]{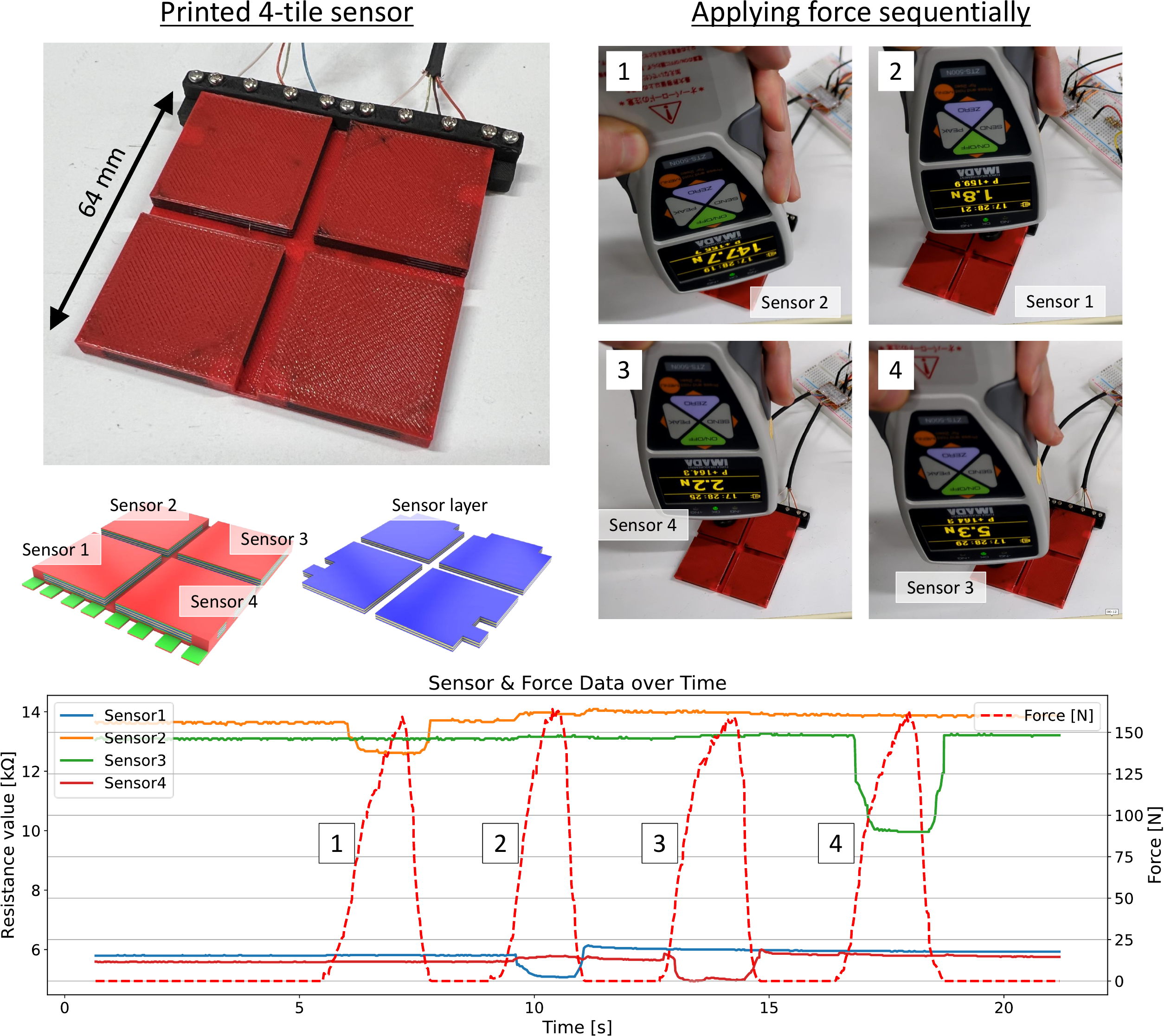}
  \caption{
  4タイルのセンサについて、力を加えた際の抵抗値変化を示す。
  }
  \label{4tile}
\end{figure}

\subsection{センサの応用}
センサの応用として、足裏センサとしての利用、ロボットハンドの一体印刷および
認識動作を行った。
\subsubsection{足裏センサ}
足裏センサとしての利用を\figref{sole}に示す。
この実験では、足裏の形状で6タイルのセンサを作成し、
足裏に装着して歩行、階段を登る動作を行い、
センサの値のパターン変化を測定した。
タイルごとに抵抗値が大きく異なるために、
パターンの比較のため、
Arduinoで直接した、
分圧回路の電圧の値をプロットしている。
歩行動作では、全てのセンサにおいて、歩行リズムに合わせた
規則的な値の変化がみられる。
一方、階段においては、かかと側にある5,6番のセンサの値がほぼ変わらず、
他の4つの値の規則的変化がみられる。
このセンサでも、人間の動作に応じたパターン変化が十分に得られている。
このように、提案したセンサが人体動作計測においても有用であることが示された。
\begin{figure}[t]
  \centering
  \includegraphics[width=1.0\columnwidth]{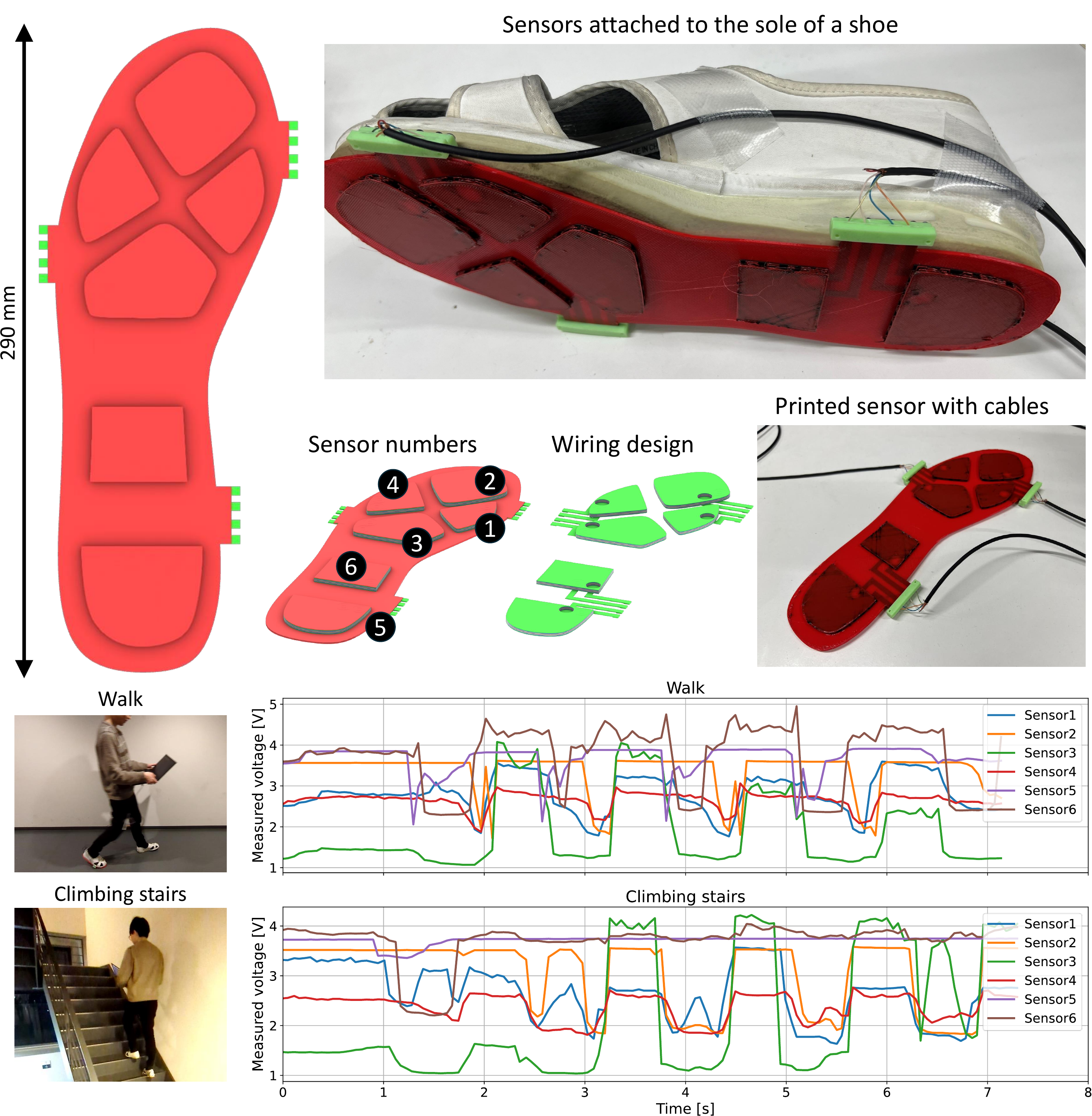}
  \caption{
  足裏のセンサへの応用を示す。
  図の上部には、設計・印刷したセンサおよび
  靴の裏への取り付けを示す。
  下部には、歩行および階段登り時のセンサ値の変化を示す。
  }
  \label{sole}
\end{figure}

\subsubsection{ロボットハンド}
ロボットハンドへの触覚利用を\figref{pr2}に示す。
この実験では、双腕ロボットであるPR2の右手に、4タイルのセンサを取り付けて、
物体を持つ動作および認識動作を行った。
センサは、ハンドとの接続部としてPLAフィラメントを含めた一体印刷で作成し、
感度を高めるためにセンサ表面に突起を設けている。
\figref{pr2}の中部には、4つの異なる物体を、ハンドの
先端側から奥側で持ち、対応するセンサの抵抗が1から2k $\Omega$程度変化する
様子を示す。
なお、タイルごとの抵抗値の違いは、配線部分の抵抗値の違いによるものであると
考えられる。
次に、\figref{pr2}の下部には、物体把持時のセンサ値の変化が、手の奥側か先端側かに
応じて、ハンドの位置を調整して物体を筒へと入れる動作を行った。
最初はセンサ2の値が変化したために先端側の位置に物体があるとして位置を決定し、
2回目はセンサ4の値が変化したために手の奥側に物体があるとして位置を決定した。
この実験においては、このセンサを利用したロボットへの応用可能性だけでなく、
3Dプリンタによる一体成型を通した、センサの簡単な組み込みが可能であることを示した。

これまでの4つの実験を通して、
センサの特性の測定および、
構造やパラメータ変化により特性を変化させられること、
複数タイルセンサとして人体計測やロボットへの応用が可能であること、
一体成型による組み込みが容易であることが示された。
\begin{figure}[t]
  \centering
  \includegraphics[width=1.0\columnwidth]{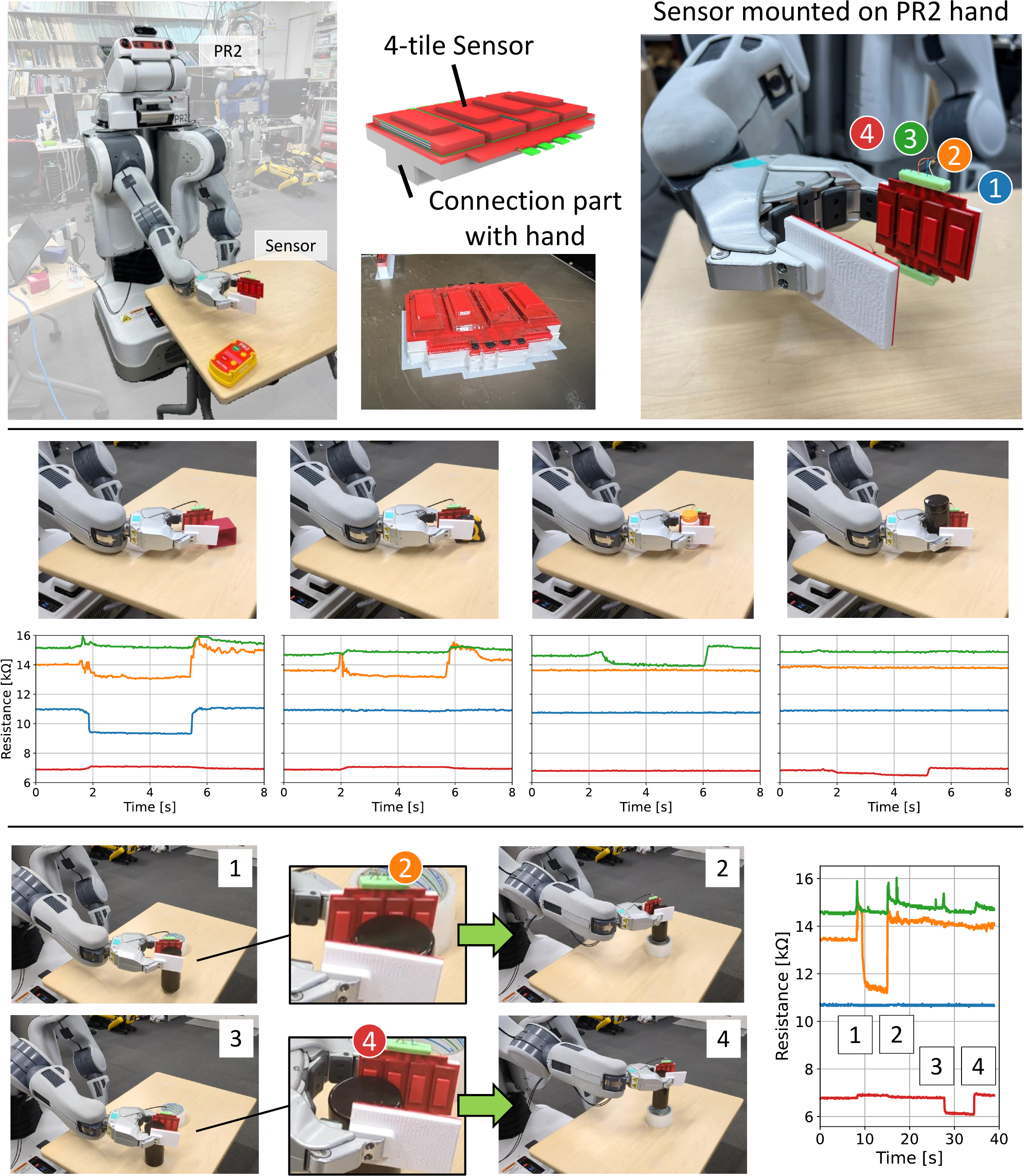}
  \caption{
  ロボットハンドへの組み込みと認識動作を示す。
  図の上部では、導電性・非導電性TPU・PLAによる一体印刷で
  作られたセンサと、PR2の右手への取り付けを示す。
  図の中央では、4つの異なる物体を持ち、センサ値の変化を示す。
  図の下部では、物体の位置をセンサより認識し、筒へと入れる動作を示す。
  }
  \label{pr2}
\end{figure}
}

\section{Conclusion}\label{sec:conclusion}
\switchlanguage%
{%
In this study, we proposed a tactile sensor utilizing the infill patterns of layer-based 3D printing. By leveraging the
hierarchical structure formed through sparse infill using conductive and non-conductive flexible filaments, we
demonstrated that changes in pressure result in corresponding variations in resistance. Through experiments, we showed
that the sensor's characteristics can be adjusted by modifying its structure and parameters, that multi-tile sensors can
be applied to human body measurements and robotic recognition, and that easy sensor integration is possible through
single-step multi-material integrated printing. 
% These results indicate that M3D-skin serves as a novel
% approach to tactile sensing that is simple, adaptable to arbitrary shapes, and well-suited for embedded applications.
These results indicate that the proposed method offers a simple, shape-adaptable, and embedded-suitable approach to tactile sensing.

Although the proposed sensor demonstrated promising performance, 
challenges remain in terms of durability, hysteresis handling, and improving the sensitivity range.
Future prospects include developing a design methodology that considers flexibility and sensitivity range as design
requirements, integrating sensors with robotic components through single-step multi-material printing with a larger
number of sensor tiles, and achieving more accurate pressure measurement through further analysis of sensor data.
}%
{%
本研究では、積層型3Dプリントのインフィルパターンを活用した
触覚センサを提案した。
導電性および非導電性の柔軟フィラメントの、疎なインフィルによる
階層構造を利用して、
圧力変化に応じた抵抗値変化が得られることを示した。
実験を通して、
構造やパラメータ変化により特性を変化させられること、
複数タイルセンサとして人体計測やロボットでの認識といった応用が可能であること、
一体成型による容易なセンサ組み込みが可能であることを示した。
これらの結果から、本手法が、
簡単かつ任意形状として使用でき、組み込み応用にも適する
触覚センシングの新たな手法として有用であることを示した。
今後の展望として、柔軟性や感度幅を設計要件として考慮した設計手法や、
より多いタイル数のセンサを用いた、ロボットの部品との一体印刷、
センサ値のさらなる解析に基づく正確な圧力計測実現が挙げられる。
}

{
  %\footnotesize
  %\small
  %\bibliographystyle{junsrt}
  \bibliographystyle{IEEEtran}
  \bibliography{bib}
}

\end{document}